\pdfoutput=1

\documentclass[11pt]{article}

\usepackage[preprint]{acl}

\usepackage{times}
\usepackage{latexsym}

\usepackage[T1]{fontenc}

\usepackage[utf8]{inputenc}

\usepackage{microtype}

\usepackage{inconsolata}

\usepackage{graphicx}
\usepackage{booktabs}
\usepackage{multirow}

\usepackage{dirtytalk} 

\usepackage{booktabs}
\usepackage{multirow}
\usepackage{adjustbox}
\usepackage{lscape}

\usepackage{xcolor}

\usepackage{tcolorbox}
\usepackage{float}
\usepackage{newfloat}
\usepackage[skip=2pt]{caption}

\usepackage{algorithm}
\usepackage{algorithmic}
\usepackage{amsmath}

\usepackage{enumitem}
\usepackage{fontawesome5}

\DeclareFloatingEnvironment[
    fileext=lob,
    listname=List of Text Boxes,
    name=Prompt,                       
    placement=htbp,
]{textbox}

\definecolor{lightgreen}{HTML}{e8f4ea}
\definecolor{darkgreen}{HTML}{b8d8be}

\usepackage{hyperref}

\title{A Structured Clustering Approach for Inducing Media Narratives}

\author{
   Rohan Das$^{1}$ \,\,
   Advait Deshmukh$^{1}$ \,\,
   Alexandria Leto$^{1}$ \,\,
   Zohar Naaman$^{1}$ \\
   \textbf{I-Ta Lee}$^{2}$ \,\,
   \textbf{Maria Leonor Pacheco}$^{1}$ \\
   $^{1}$University of Colorado Boulder \,\, $^2$Independent Researcher\\
   \texttt{\{rohan.das, maria.pacheco\}@colorado.edu} \\
}

\begin{document}
\maketitle
\begin{abstract}
Media narratives wield tremendous power in shaping public opinion, yet computational approaches struggle to capture the nuanced storytelling structures that communication theory emphasizes as central to how meaning is constructed. Existing approaches either miss subtle narrative patterns through coarse-grained analysis or require domain-specific taxonomies that limit scalability. To bridge this gap, we present a framework for inducing rich narrative schemas by jointly modeling events and characters via structured clustering. Our approach produces explainable narrative schemas that align with established framing theory while scaling to large corpora without exhaustive manual annotation.
\end{abstract}
\begin{center}
\faGithub~\href{https://github.com/blast-cu/structured-clustering-for-narratives}{Code}
\end{center}

\section{Introduction}
Media communicators actively shape the understanding of their audience by transforming multifaceted problems into digestible narratives and focusing on certain dimensions while obscuring others \citep{Entman2003CascadingAC}. This selective presentation creates interpretative frameworks that both communicators and consumers rely on to navigate complex information landscapes \citep{Pan1993FramingAA}. The resulting narratives do not simply describe reality; they actively construct it by directing attention to particular aspects of an issue while rendering others invisible \citep{Adams1979FrameAA}.

The study of media narratives reveals more than how messages are constructed; it exposes underlying power structures and shows how public opinion takes shape. When communicators use storytelling techniques such as character roles \citep{mendelsohn-etal-2021-modeling} and conflict resolution constructs \citep{frermann-etal-2023-conflicts}, they turn complicated issues into simple narratives that guide the reader toward specific conclusions \citep{Crow2016MediaIT}. This storytelling approach becomes particularly powerful when different groups compete to define controversial topics. Analysis of these communication strategies can help researchers understand how social movements build support, how public debates develop, and how harmful stereotypes become reinforced or challenged.

\begin{figure}[t]
    \centering
    \includegraphics[width=0.8\columnwidth]{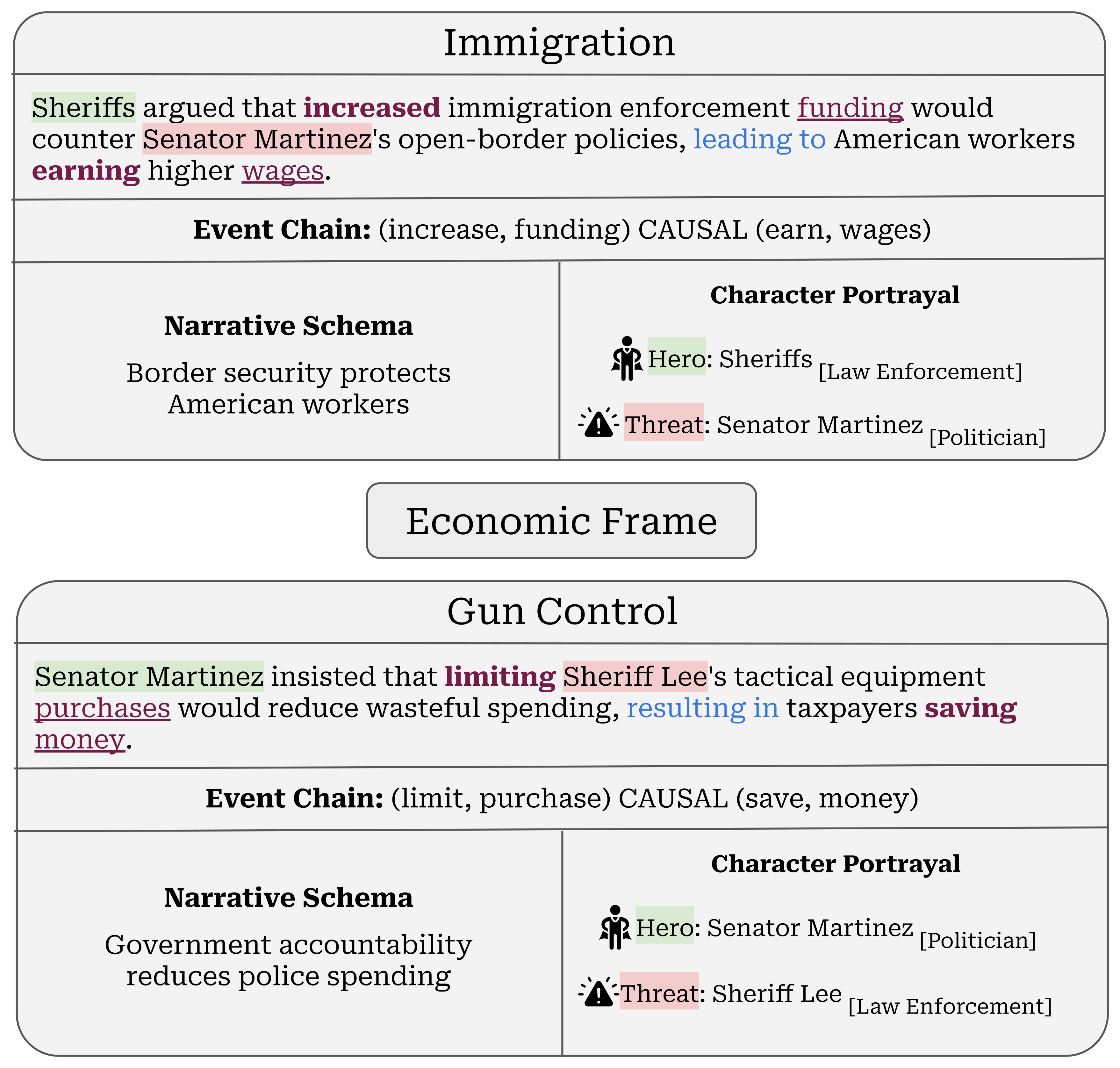}
    \caption{Law enforcement spending framed as worker protection versus government waste across policy domains. Verbs are in  \textcolor[HTML]{965575}{\textbf{bold}}. Objects are \textcolor[HTML]{965575}{\underline{underlined}}. (Verb, Object) tuples make up events. Causal markers are in \textcolor[HTML]{467ed9}{blue}. Characters portrayed as heroes are highlighted in \colorbox[HTML]{d9ead3}{green}, while threats are highlighted in \colorbox[HTML]{f4cccc}{pink}. Character groups are denoted in subscripts. The Economic frame is from \citealp{Boydstun2014TrackingTD}'s Policy Framework.}
    \label{fig:motivating_example}
\end{figure}

Natural language processing (NLP) approaches to media analysis often focus on broad and coarse-grained labels such as partisanship (e.g. left, right) \citep{Fulgoni2016AnEE} and topical media frames (e.g. political, economic, security and defense)~\citep{card-etal-2015-media}, or domain-specific taxonomies tailored to particular policy issues such as immigration \citep{mendelsohn-etal-2021-modeling} and the economy \citep{leto-etal-2024-framing}. While coarse-grained labels enable scalable annotation across domains, they tend to overlook nuanced storytelling and thematic structures emphasized in communication research \citep{Entman2003CascadingAC, Fairhurst2005ReframingTA}. In contrast, domain-specific taxonomies capture ideological and cultural subtleties but lack the ability to generalize to other domains. This fragmentation limits consistent narrative analysis across domains; bridging these perspectives requires methods that balance scalability with interpretative depth, supporting rich, meaningful media analysis aligned with how narratives function in public discourse.

In this paper, we present a framework that aims to balance these two objectives (see Fig. \ref{fig:motivating_example}). We start with a bottom-up approach by identifying \textbf{events} and their \textbf{causal relations} in text to form \textbf{event chains}. We augment these chains with character-level information. That is, we identify their \textbf{narrative roles} and assign them to \textbf{character groups}. Subsequently, we \textbf{cluster similar narrative chains} using these character-role assignments as constraints. These role-based constraints enable the separation of superficially similar narratives by revealing the underlying framing distinctions. This clustering process results in cohesive and semantically nuanced \textbf{narrative schemas} that transcend surface-level topical similarity.  

We argue that inducing these structured narrative schemas is a way to bring together the notion of narratives from both a communication theory and an NLP perspective. Communications scholarship treats a narrative as a culturally shared sense-making schema that guides how audiences interpret events \citep{Fisher1985TheNP}.
NLP approaches invert this perspective, building narratives bottom-up from extractable atomic units. For example, by modeling events using predicate-argument structures~\citep{chambers-jurafsky-2008-unsupervised,roth-lapata-2015-context,lee-etal-2020-weakly}, and characters through embeddings \citep{Kim2018UnderstandingAA} or  high-level descriptive summaries \cite{fokkens-etal-2018-studying, brahman-chaturvedi-2020-modeling,gurung-lapata-2024-chiron}. While scalable, these pipelines reduce stories to verbs, roles, and entities, seldom capturing the evaluative and ideological dimensions central to communication theory. In contrast, we present a unified approach that employs both event- and character-centric modeling to construct high-level narrative schemas that capture broader framing dimensions. This framework enables us to induce narrative schemas at scale from arbitrarily large corpora in a domain-agnostic manner, producing structures that align with the conflict, evaluation, and resolution patterns identified in media studies~\cite{Entman1993FramingTC}.
\section{Related Work}

\begin{figure*}[ht!]
    \centering
    \includegraphics[width=0.8\textwidth]{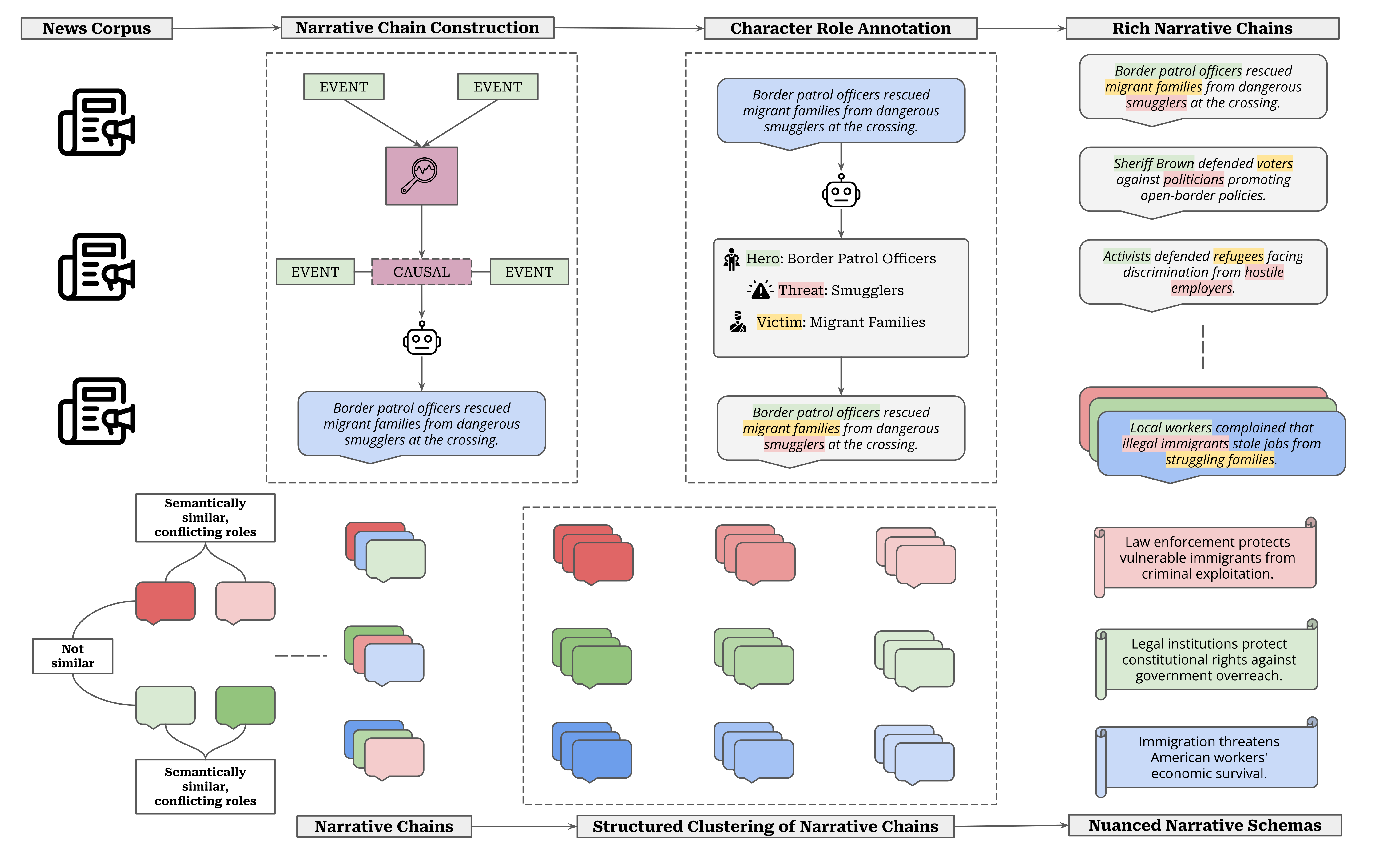}
    \caption{For a large scale news corpus, we first construct narrative event chains and obtain character and role annotations for them. We then cluster these narrative chains using the character and role information as constraints. The generated narrative clusters are representative of fine-grained and nuanced narrative schemas.}
    \label{fig:framework}
\end{figure*}

\paragraph{NLP Approaches to Narrative Modeling.} \citealp{chambers-jurafsky-2008-unsupervised}'s seminal work on narrative schemas introduced unsupervised induction of narrative templates, capturing coherent sequences of typed event chains from raw text. This conceptualization of narratives gave way to neural methods that learned prototypical event sequences directly from tokenized text using different variations of recurrent neural networks~\cite{pichotta-mooney-2016-using}, and multi-relational graph approaches where nodes represent events interconnected through typed edges reflecting temporal and causal relations~\citep{lee-goldwasser-2019-multi,lee-etal-2020-weakly}. Complementing structural approaches, \citealp{antoniak-etal-2024-people} framed the task of story detection as identifying sequences of events involving one or more entities, using theory-driven features such as event presence and verb tense to detect narratives across online communities.

Character-centric modeling has also advanced through data-driven benchmarks and neural architectures. \citealp{brahman-etal-2021-characters-tell}’s LiSCU dataset established character identification and description generation tasks from literary summaries, triggering neural models that recover persona traits, inter-character relations, and emotional trajectories~\cite{brahman-chaturvedi-2020-modeling}. Recent character-sheet frameworks use large language models (LLMs) for extracting question-answer pairs and performing entailment validation to populate structured persona profiles - detailing goals, background, dialogue style, and inferred motivations - enhancing accuracy in masked character prediction tasks \citep{gurung-lapata-2024-chiron}. 

\paragraph{Narrative Analysis for Media Framing.}
In the last decade, most of the NLP work in media framing has conceptualized frames as coarse-level topics~\cite{Boydstun2014TrackingTD, card-etal-2015-media,naderi-hirst-2017-classifying, ji-smith-2017-neural, field-etal-2018-framing,  khanehzar-etal-2019-modeling}. More recently, however, there has been increasing interest in uncovering narratives and narrative devices used in news articles.  Many works have focused on analyzing the characterization of various entities \citep{card-etal-2016-analyzing, ziems-yang-2021-protect-serve, frermann-etal-2023-conflicts}. Others focus on automatically identifying events and the relationships between them \cite{zhao-etal-2024-media, das-etal-2024-media}. Practitioners have also included entity information with additional article-level frame information for finer-grained analysis \cite{otmakhova-frermann-2025-narrative}. To our knowledge, this is the first computational work to combine detailed character-level information with event-level information for a more comprehensive view of media narratives.

\section{Inducing Narrative Schemas}

\begin{table*}[ht!]
\centering
\resizebox{0.9\textwidth}{!}{%
\begin{tabular}{ll}
\hline
\textbf{Causal Narrative Chain}                         & \textbf{Verbalization and Character Roles}                                                                                          \\ \hline
\texttt{((withhold, permit), CAUSAL, (introduce, law))} & \begin{tabular}[l]{@{}l@{}}\colorbox[HTML]{d9ead3}{Mayor Lou Barletta}\textsubscript{[POLITICIANS]} introduced a law that withholds permits \\ from companies hiring \colorbox[HTML]{f4cccc}{illegal immigrants}\textsubscript{[IMMIGRANTS]} in response to \\ concerns over illegal immigration in Hazleton.\end{tabular}   \\

\hline
\texttt{((exempt, police), CAUSAL, (turn, police))} & \begin{tabular}[l]{@{}l@{}}The New York law's failure to exempt \colorbox[HTML]{ffe599}{police}\textsubscript{[LAW ENFORCEMENT]} from the \\ magazine ban caused them to become criminals when carrying standard \\ firearms, as noted by \colorbox[HTML]{d9ead3}{Mayor Michael Bloomberg}\textsubscript{[POLITICIANS]}.\end{tabular}                                            
\\ \hline
\end{tabular}%
}
\caption{Examples of causal narrative chain verbalizations and their character role annotations. Characters portrayed as heroes are highlighted in \colorbox[HTML]{d9ead3}{green}, threats are highlighted in \colorbox[HTML]{f4cccc}{pink}, and victims are highlighted in \colorbox[HTML]{ffe599}{yellow}. Character groups are denoted in subscripts.}
\label{tab:llm-expansion}
\end{table*}

We present a multi-stage pipeline for inducing narrative schemas from large news corpora (Fig. \ref{fig:framework}).

\subsection{Task Definition}

\textbf{Narrative schemas} represent a fundamental concept in computational narrative understanding that goes beyond simply organizing story elements. While event schemas primarily focus on extracting sequences of actions and their temporal relationships~\citep{chambers-jurafsky-2008-unsupervised, chambers-jurafsky-2009-unsupervised}, narrative schemas function as information compression systems that encompass the broader evaluative and social dimensions determining why certain stories gain cultural significance, influence specific communities, and serve as vehicles for social understanding~\citep{piper-2023-computational}.

We define the computational task of narrative schema induction as the automatic discovery and extraction of culturally and socially important narrative schemas from large-scale, domain-agnostic text corpora with minimal manual annotation or supervision.

\subsection{Constructing Narrative Chains}

We systematically construct narrative chains by extracting events and causal relationships between them, and generating rich textual representations that capture these relations in context. 

\paragraph{Extracting Events.} We follow the convention of representing an event as a tuple of predicate tokens (verb) and their dependencies (object) \citep{GranrothWilding2016WhatHN}. To extract event mentions from text, we use the ETypeClus framework \citep{shen-etal-2021-corpus}. To find verb and object heads in sentences, we use an off-the-shelf dependency parser\footnote{We use the Spacy \texttt{en\_core\_web\_lg} model.} and select all non-auxiliary verb tokens.
Next, we find the corresponding object head for each candidate verb based on whether the sentence is in active or passive voice. We then process the entire corpus to extract a list of all $\texttt{event} \Rightarrow (\texttt{verb}, \texttt{object})$ mentions in each article.

\paragraph{Extracting Causal Relations.} We adapt the approach proposed in DAPrompt to extract causally related pairs of events in each article~\cite{Xiang2023DAPromptDA}. The method works by first assuming a causal relationship between a pair of events and then evaluating the validity of the assumption via prompting. The model receives as input the sentences that contain events, their surrounding context, and the assumption template. We provide up to 50 tokens left and right of each event sentence.

To obtain training labels for our target datasets, we prompt an instruction-tuned Llama 3.3 70B model \citep{Dubey2024TheL3} on 20,000 sampled event pairs. We distill this knowledge by training DAPrompt on the resulting silver standard labels, enabling efficient inference while approximating the teacher model's performance. We restrict inference to event pairs within two sentences. This step yields \texttt{(event, causal, event)} triplets that capture the causal relationships within each article. Prompts are included in App. \ref{box:causal_prompt} and \ref{box:causal_prompt_ctd}. 

\paragraph{Verbalizing Causal Narrative Chains.} To transform extracted triplets into rich textual representations, we use the DeepSeek-R1-Distill-Llama-70B reasoning model in a zero-shot setting~\citep{DeepSeekAI2025DeepSeekR1IR}. We provide the model with the source article along with its domain and the corresponding narrative chain triplet. The model is instructed to generate concise sentences that coherently encapsulate the underlying causal event sequences. The model is then instructed to incorporate all relevant characters who participate directly in the narrative chain. The prompts are included in App. \ref{box:chain_verb_prompt}. Examples of verbalizations are shown in Tab. \ref{tab:llm-expansion}.

\subsection{Extracting Characters and Roles}
To enhance our narrative representations, we extract the characters relevant to each narrative chain and determine their respective narrative roles in the context of the full article.

\paragraph{Identifying Character Groups.} To identify all salient character groups in a corpus, we prompt instruction fine-tuned Llama 3.3 70B \cite{Dubey2024TheL3} in a five-shot setting to identify all mentions of main characters in a given news article. Then, we cluster mentions (e.g. \say{undocumented minors}) to obtain relevant character groups (e.g. \say{immigrants}). To do so, we compile a list of expected character groups for a given domain based on the list of extracted character mentions. For immigration, for example, this could include ``politicians'' and ``law enforcement''. These expected groups serve as initial centers for the k-means clustering algorithm. We then use k-means to divide the list of characters into a fixed number of clusters. Finally, we manually label the character clusters to obtain the final list of character groups present in the given corpus for each domain. Note that this manual annotation process is quick, as we only need to provide labels at the cluster level. See App. \ref{app:char_ext_details} for details. Prompts are included in App. \ref{box:char_ext_prompt}. 

\vspace{-0.1mm}

\paragraph{Assigning Character Roles.} We recognize that effective narrative analysis requires the systematic characterization of characters according to their archetypal roles within causal sequences. We adopt \citealp{Shanahan2011NarrativePF}'s narrative policy framework and explicitly model \emph{heroes} (agents that drive positive outcomes), \emph{threats} (characters that create conflict or harm), and \emph{victims} (those who suffer consequences) as a means of understanding the structure and meaning of the narrative. Identifying these roles is essential for modeling narratives, as it captures the moral and functional positioning of actors within our causal sequences, as shown in Fig. \ref{fig:motivating_example}. 

To identify all characters and their roles, we draw inspiration from \citep{stammbach-etal-2022-heroes} and prompt the DeepSeek-R1-Distill-Llama-70B reasoning model in a zero-shot setting~\citep{DeepSeekAI2025DeepSeekR1IR}.  We provide the model with the source article along with its domain and the corresponding narrative chain verbalization. We also provide the list of relevant character groups that were extracted in the previous step as well as the descriptions of the hero/threat/victim roles in the context of the source domain. The LLM is instructed to extract all relevant characters from the narrative chain, tag them with their appropriate character group, and identify the role they play in the narrative (Hero/Threat/Victim). It also identifies the general stance of the chain with respect to the relevant policy domain (e.g., support/oppose immigration). Prompts are included in App. \ref{box:char_role_prompt} and \ref{box:char_role_prompt_ctd}. Examples of role annotations are shown in Tab. \ref{tab:llm-expansion}.

\subsection{Structured Clustering}
While character role identification establishes the narrative functions and agency of actors within individual narrative chains, our broader objective requires systematic organization of these chains to reveal the underlying thematic patterns across the corpus. Traditional clustering approaches based solely on textual similarity may inadvertently group narratives that, despite surface-level resemblance, convey fundamentally different moral or thematic messages due to contrasting character role configurations. To address this challenge, we employ a structured clustering process that incorporates character and role information as additional constraints, thereby ensuring that narratives with divergent character dynamics are appropriately differentiated. This approach enables the direct induction of coherent schemas from the resulting narrative clusters, where each cluster represents a distinct thematic category grounded in both causal event structure and character role consistency.

\paragraph{Constraints.} We generate cannot-link constraints (see App. \ref{app:constraints}) for all pairwise combinations of chains across the corpus whenever their character groups and stances exhibit conflicting configurations (different roles for the same character group, different stances). Illustrative examples are presented in Tab. \ref{tab:constraints_examples}.

\begin{table}[]
\resizebox{\columnwidth}{!}{%
\begin{tabular}{@{}lll@{}}
\toprule
\textbf{Canonical Form of Chain A}                                                                 & \textbf{Canonical Form of Chain B}                                                                                      & \textbf{Constraint} \\ \midrule
\begin{tabular}[c]{@{}l@{}}Immigrants: Victim\\ Law Enforcement: Threat\\ Stance: Pro\end{tabular} & \begin{tabular}[c]{@{}l@{}}Immigrants: Threat \\ Law Enforcement: Hero\\ Stance: Pro\end{tabular}                       & Cannot-Link         \\ \midrule
\begin{tabular}[c]{@{}l@{}}Government: Threat\\ Politicians: Hero\\ Stance: Anti\end{tabular}       & \begin{tabular}[c]{@{}l@{}}Gun Rights Advocates: Hero, \\ Politicians: Hero\\ Gun Control Advocates: Threat\end{tabular} & May-Link         \\ \bottomrule
\end{tabular}%
}
\caption{Illustrations of canonical representations of character-role configurations of narrative chains. We show cases where chains cannot be linked and may be linked. We only consider cannot-link constraints. See App. \ref{app:constraints} for more details.}
\label{tab:constraints_examples}
\end{table}

\paragraph{Constrained Clustering Objective.} Drawing inspiration from \citealp{Basu2004ActiveSF}'s pairwise constrained clustering framework, we propose a k-means-based clustering algorithm that only considers cannot-link constraints. Let $\mathcal{X} = \{x_i\}^n_{i=1}$ be the set of narrative chains and $k$ be the number of clusters. The goal is to find a disjoint $k$-partitioning $\{\mathcal{X}_h\}^k_{h=1}$ (with each partition having a centroid $\mu_h$), such that the total distance between the chains and the cluster centroids is locally minimized. 

Let $\mathcal{C}$ be the set of cannot-link constraints such that $(x_i,x_j) \in \mathcal{C}$, implies $x_i$ and $x_j$ should be assigned to different clusters. The tuples in $\mathcal{C}$ are order-independent, i.e., $(x_i,x_j) \in \mathcal{C} \Rightarrow (x_j,x_i) \in \mathcal{C}$. Let $w_{c}$ be the weight of the constraints. Let $l_i$ be the cluster assignment of a chain $x_i$, where $l_i \in \{h\}^k_{h=1}$. The cost of violating a cannot-link constraint $(x_i, x_j) \in \mathcal{C}$ is given by $w_{c}\mathbf{1}[l_i = l_j]$, i.e., if chains violating these constraints are assigned to the same cluster, the incurred cost is $w_{c}$. Here, $\mathbf{1}$ is the indicator function, with $\mathbf{1}[true]=1$ and $\mathbf{1}[false]=0$. Thus, the constrained clustering problem can be formulated as minimizing the following objective function, where the chain $x_i$ is assigned to the cluster $\mathcal{X}_{l_{i}}$, with centroid $\mu_{l_{i}}$:

\vspace{-5mm}

\begin{equation}
\small
\mathcal{R} = \frac{1}{2} \sum_{x_i \in \mathcal{X}} \|x_i - \mu_{l_i}\|^2 
+ \sum_{(x_i, x_j) \in C} w_{c} \mathbf{1}[l_i = l_j] 
\end{equation}

\vspace{-3mm}

\paragraph{Constraint Aware Centroid Initialization.} \citealp{Basu2004ActiveSF} demonstrate that centroid initialization in centroid-based algorithms such as k-means significantly improves when incorporating supervision through pairwise point constraints. Following this insight, our initialization procedure seeks to obtain robust estimates of cluster centroids by implementing a constraint-aware variant of the k-means++ initialization algorithm~\citep{Arthur2007kmeansTA} . Let $\mathcal{S} = \{\mu_{s_1}, \mu_{s_2}, \ldots, \mu_{s_m}\}$ denote the set of already selected centroids in the current step, where $m \in \{1, 2, \ldots, k-1\}$ and each $\mu_{s_j} = x_{s_j}$ corresponds to a selected data point. For each candidate chain $x_i \in \mathcal{X}$, the selection maximizes:

\vspace{-5mm}

\begin{equation}
\small
\mathcal{O}(x_i) = \min_{\mu_s \in \mathcal{S}} \|x_i - \mu_s\|^2 + w_{c} \sum_{\mu_s \in \mathcal{S}} \mathbf{1}[(x_i, \mu_s) \in \mathcal{C}]
\end{equation}

\vspace{-3mm}

\noindent where the first term represents the squared distance from the candidate chain $x_i$ to its nearest already selected centroid, and the constraint bonus term encourages the selection of chains that cannot be linked with already existing centroids. This approach promotes constraint satisfaction during initialization by preferentially selecting centroids that are restricted to be in different clusters, thereby providing a more informed starting configuration for the subsequent clustering algorithm.

\paragraph{Clustering Algorithm.} An SBERT model \citep{reimers-gurevych-2019-sentence} that
was trained for semantic search tasks is used to compute sentence embeddings for each of the chain verbalizations and is provided as input to our clustering algorithm. The algorithm (see App. \ref{app:cluster_algo}) alternates between cluster assignment and centroid re-estimation steps. In the assignment step, each chain $x_i \in \mathcal{X}$ is assigned to the cluster that minimizes the sum of its distance to the cluster centroid and the cost of constraint violations incurred by that assignment. The centroid re-estimation step follows standard k-means, where each cluster centroid $\mu_h$ is computed as the mean of its assigned chains.

\subsection{Narrative Schema Attribution} Following the clustering process, we employ automatic schema attribution using LLMs to characterize the narrative content of the clusters. We sample narrative chains and their character role information from each cluster using a two-stage approach detailed in App. \ref{app:cluster_sampling_strategy}. These are provided as input to the DeepSeek-R1-Distill-Llama-70B reasoning model in a zero-shot setting~\citep{DeepSeekAI2025DeepSeekR1IR}. The model is instructed to generate a schema that best represents the cluster's overarching narrative and to identify and include \citealp{Entman1993FramingTC}'s \textit{issue}, \textit{evaluation}, and \textit{resolution} framing elements when detected. Prompts are included in App. \ref{box:theme_attr_prompt} and \ref{box:theme_attr_prompt_contd}.

\section{Evaluation}

\begin{table*}[ht!]
\resizebox{\textwidth}{!}{%
\begin{tabular}{lccccc|ccc}
\hline
\multicolumn{1}{c}{\multirow{2}{*}{\textbf{Domain}}} & \multirow{2}{*}{\textbf{Clusters}} & \multirow{2}{*}{\textbf{Model}}                                  & \multicolumn{3}{c|}{\textbf{Top 25\% Narrative Chains}}                                   & \multicolumn{3}{c}{\textbf{All Narrative Chains}}                                         \\ \cline{4-9} 
\multicolumn{1}{c}{}                                 &                                    &                                                                  & \textbf{Frame F1} & \textbf{Exact Match Purity} & \textbf{Avg. Role Purity} & \textbf{Frame F1} & \textbf{Exact Match Purity} & \textbf{Avg. Role Purity} \\ \hline
\multirow{2}{*}{Immigration}                         & \multirow{2}{*}{150}               & \begin{tabular}[c]{@{}c@{}}k-means\end{tabular}      & 33.22                        & 32.31                       & 86.88                        & 41.19                        & 26.90                        & 80.79                        \\ \cline{3-3}
                                                     &                                    & \begin{tabular}[c]{@{}c@{}}Structured \\ Clustering\end{tabular} & \textbf{36.96}               & \textbf{37.83}              & \textbf{87.77}               & \textbf{42.32}               & \textbf{32.79}              & \textbf{81.48}               \\ \cline{2-9} 
\multirow{2}{*}{Gun Control}                         & \multirow{2}{*}{200}               & \begin{tabular}[c]{@{}c@{}}k-means\end{tabular}      & 32.86                        & 35.16                       & 86.40                        & 37.65                        & 29.22                       & 81.18                        \\ \cline{3-3}
                                                     &                                    & \begin{tabular}[c]{@{}c@{}}Structured \\ Clustering\end{tabular} & \textbf{36.45}               & \textbf{42.71}              & \textbf{89.24}               & \textbf{41.68}               & \textbf{36.66}              & \textbf{82.90}               \\ \hline
\end{tabular}%
}
\caption{Performance evaluation across three metrics: frame prediction F1-score, exact match purity, and overall role purity. Metrics are computed for all narrative chains and for the top 25\% chains closest to cluster centroids. Results compare our method against k-means baseline on immigration and gun control domains. Number of clusters for each domain was selected based on the process outlined in App. \ref{app:best-k}. Bold values indicate superior performance.}
\label{tab:clustering-stats}
\end{table*}

\subsection{Dataset} We collect news articles from two distinct policy domains: immigration and gun control. We used the Media Frames Corpus \citep{card-etal-2015-media}, an annotated news dataset with 15 distinct framing categories. We sampled 2,000 documents from each domain and induced narratives using our framework. Data statistics and pre-processing steps are reported in App. \ref{app:dataset}.

\subsection{Narrative Chains} 

\paragraph{Causal Relations.} We curated and annotated a development set from the immigration and gun control datasets. Two independent annotators labeled 200 event pairs for each dataset. A third annotator broke ties. The average Krippendorff's-$\alpha$ score of 0.76 (0.71 for Immigration and 0.81 for Gun Control) indicates good inter-annotator agreement. DAPrompt trained on silver labels achieves an F1 score of $ 58.46 (\pm 2.06) $. The teacher model achieved an F1 score of 68.18, suggesting that we maintain 85\% of teacher performance. This is comparable to recent causal relation prediction methods \cite{Cheng2024ASO} and the high recall ($77.9 \pm 3.1$) is well suited to maximize coverage. Further details are provided in App.~\ref{app:causal-eval}.

\paragraph{Narrative Chain Verbalizations.} We sampled 50 narrative chains from each domain for human evaluation. Two independent annotators assessed chain verbalizations using a 4-point scale, assigning 0/1 values to four criteria: completeness of events, completeness of entities, semantic fidelity to the underlying causal sequence, and absence of hallucinations. The evaluation yielded mean scores of 3.3 out of 4 for immigration and 3.49 out of 4 for gun control. Inter-annotator reliability was excellent for gun control (ICC(2,1) = 0.90) and good for immigration (ICC(2,1) = 0.76)\footnote{Immigration = 95\% CI [0.84, 0.94], Gun Control = 95\% CI [0.62, 0.86], p < 0.001}.

This evaluation revealed an effective capture of events and entities without hallucinations, although fidelity suffered due to occasional oversimplification of causal connections, likely resulting from errors in causal predictions. No egregiously incorrect verbalizations were observed in either dataset.

\paragraph{Character and Role Annotations.} Annotators also evaluated character, role, and stance annotations using a 5-point scale, assigning 0/1 values to five criteria: completeness of entity identification, accuracy of character group classification, correctness of role assignment, stance identification, and absence of hallucinations. The system achieved strong performance with mean scores of 4.0 out of 5 for immigration and 4.73 out of 5 for gun control. Inter-annotator reliability was excellent for gun control (ICC(2,1) = 0.83) and good for immigration (ICC(2,1) = 0.79)\footnote{Immigration = 95\% CI [0.72, 0.90], Gun Control = 95\% CI [0.66, 0.87], p < 0.001}.

\subsection{Narrative Clustering}

\paragraph{Cluster Purity.}
We use two complementary purity measures that quantify the homogeneity of character-role configurations within clusters. \textit{Exact match purity} assesses clustering quality by considering the complete canonical character-role configuration of each narrative chain, where each configuration represents a complete mapping of character groups to narrative roles (e.g., {Immigrants: Victim, Law Enforcement: Threat, Politicians: Hero}). This metric measures how many narrative chains within each cluster share identical canonical forms. In contrast, \textit{role-based purity} decomposes each narrative chain into individual character-role combinations and measures clustering consistency at the granular level. Mathematical formulations are described in App. \ref{app:purity-math}.

\paragraph{Predictive Signal from Clusters.}
To assess the correlation between narrative clusters and high-level frames, we examine whether cluster assignments contain enough signal to discriminate between frame categories. This analysis is designed not to optimize predictive performance but rather to quantify the extent to which frame-relevant information is implicitly encoded within the discovered narrative structures. We employ a logistic regression classifier that takes cluster frequency vectors as input features, where each document is represented by the frequency counts of clusters assigned to its constituent narrative chains (App. \ref{app:feature-vector}). Performance is evaluated using F1 on a held-out development set comprising 20\% of the total dataset.

\paragraph{Results.} All metrics are reported in Tab. \ref{tab:clustering-stats}. Through systematic experimentation, we determined the optimal number of clusters for each domain (see App. \ref{app:best-k}). Comparative analysis against standard k-means clustering demonstrates consistent improvements across all evaluation metrics for both domains. The incorporation of character-role constraints yields more semantically coherent and nuanced clusters, with corresponding improvements in frame prediction accuracy (up to 3-4 points) and exact match purity (up to 5-7 points) when evaluated on the top quartile of narrative chains within each cluster. Detailed individual role purity scores are presented in App. \ref{app:role-purity}.

\paragraph{Intrusion Test.}
We further assess the quality of the clusters with an intrusion detection task. For each test instance, we select two randomly sampled narrative chains from a target cluster and introduce a randomly sampled chain from a different cluster as a distractor. Two independent annotators are tasked with identifying the intruding chain. The underlying premise is that well-formed clusters capturing coherent high-level narrative patterns should facilitate accurate intruder identification. We perform the intrusion detection task on 150 examples derived from clustering results of both standard k-means and our proposed algorithm across both datasets. We found that constrained clustering significantly improves intrusion detection for gun control (75 vs. 85). However, for immigration, we observe the opposite trend (86 vs. 83). We hypothesize that this could be due to the following reasons: 1) During hyper-parameter tuning, we select a lower constraint weight for immigration (0.01 vs. 0.1), which results in constraints being applied more loosely. 2) We systematically observe a lower IAA for immigration across all tasks, suggesting potential dataset artifacts. We leave a closer examination of this result for future work. Full accuracy and agreement metrics are in App. \ref{app:intrusion-details} (Tab. \ref{tab:intrusion-test}). 

\begin{figure*}[ht!]
    \centering
    \includegraphics[width=1.0\textwidth]{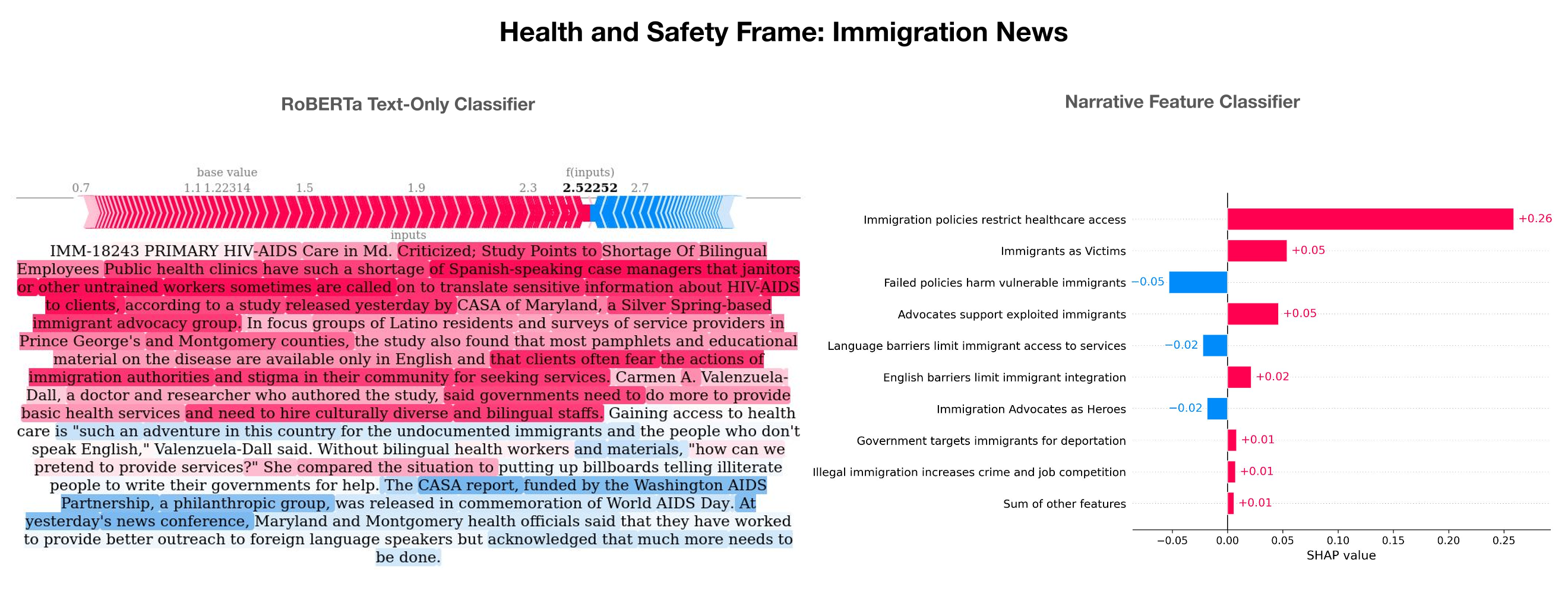}
    \caption{SHAP feature importance comparison between RoBERTa embeddings and narrative features for frame prediction on a single test instance, showing both approaches capture similar semantic patterns. Red/blue indicate positive/negative influence; deeper shades indicate higher influence. Full length narrative schemas in App. \ref{app:single-instance-themes}.}
    \label{fig:shap-individual}
\end{figure*}

\subsection{Schema Attribution Analysis}

To evaluate the quality of schema attribution, we had a trained linguist perform a qualitative analysis. This process focused on analyzing 50 resulting schemas from each dataset -- selected based on the strategy outlined in App. \ref{app:schema-analysis} -- for overall schema quality, as well as the presence of \citet{Entman1993FramingTC}'s frame elements in the schema. We find that 94\% of the schemas from the immigration set and 96\% of the schemas from the gun control set were of high-quality; these not only meet high standards of overall coherence, but also strike a balance between coverage and specificity to the sentences included in the context. Additionally, we find high rates of the \textit{evaluation} frame element in schemas. While the \textit{evaluation} element was most prominent, we also observed a relatively high number of schemas that contained all three elements simultaneously (namely \textit{issue}, \textit{evaluation}, and \textit{resolution}). We believe that the high quality of generated schemas -- as well as the presence of frame elements -- provides evidence that the results of our Narrative Schema Attribution step are useful in evaluating framing in news media. Frame element definitions and example schemas are provided in App. \ref{app:schema-analysis}. 

\subsection{Error Analysis}

We conducted a qualitative error analysis to identify failure modes in our pipeline. We categorize errors into five types, each illustrated with a detailed example in App.~\ref{app:error-analysis} (Tab.~\ref{tab:error-analysis-1}).

\paragraph{Incorrect Event Extraction.} The causal relation extraction module may identify the wrong event triggers from a sentence, leading to an incorrect event pair. Since all downstream components depend on the extracted events, this produces cascading errors in the verbalization, character annotation, and schema assignment.

\paragraph{Incorrect Causal Relation Prediction.} In some cases, the model predicts a causal relationship between events where the actual relationship is more nuanced or absent entirely. This is particularly common when the article implies a policy recommendation rather than a direct causal link, causing the LLM to fabricate a causal interpretation that is unfaithful to the source text.

\paragraph{Missing Entity from the Verbalization.} The verbalization step may omit key entities mentioned in the article, attributing actions or experiences to the wrong actor. This typically occurs when multiple entities are discussed in close proximity and the LLM conflates their roles based on the broader article context rather than the specific event pair.

\paragraph{Unfaithful Verbalization.} Even when the correct events are extracted, the verbalization may invert or distort the causal direction described in the article. The resulting narrative chain misrepresents the relationship between events, though downstream schema assignment may still capture the general thematic content of the article.

\paragraph{Incorrect Character Annotation.} The character annotation module may assign incorrect narrative roles when the model's interpretation of an entity's role differs from the role implied by the article context. Additionally, because our algorithm maps fine-grained entities to coarser character groups, semantically related but distinct entities (e.g., immigrant teachers vs.\ immigrant students) may be conflated, leading to incorrect schema assignments.

\paragraph{Recommendations.} Improving upstream event extraction and causality identification yields downstream gains, as errors here cascade through the entire pipeline. A faithfulness verification step, such as entailment checking between the verbalization and source article, could mitigate unfaithful verbalizations and missing entities. Practitioners should exercise caution with implicit causality in argumentative text, where the model may overfit to causal interpretations. Adopting finer-grained, domain-specific character taxonomies could further reduce entity conflation. Notably, despite intermediate errors, schema assignment often remains thematically faithful, suggesting the pipeline is robust for aggregate-level narrative analysis.

\section{Explainability Study}

We seek to systematically examine which narrative patterns and character configurations are most predictive of specific media frames at different levels of granularity, compare the explanatory power of our narrative features against contextualized text representations, and identify the key narrative elements that most strongly signal particular media framing strategies. To perform this analysis, we employ SHAP (SHapley Additive exPlanations), a game-theoretic framework that assigns importance scores to individual features based on their contribution to predictions \citep{Lundberg2017AUA}.

\paragraph{Instance-Level Analysis.} We trained two classifiers for comparative analysis: a text-only RoBERTa model \citep{Liu2019RoBERTaAR} and a multi-layer neural network that uses narrative cluster and character role frequencies as the only input features for policy frame prediction. We employed SHAP analysis to examine feature importance for identical test instances across both classifiers, illustrating our findings in Fig. \ref{fig:shap-individual}. The analysis reveals that the narrative cluster features capture equivalent thematic information to token-level RoBERTa embeddings, while offering more succinct information. In addition, both models rely on similar high-level patterns for \textit{Health and Safety} prediction. This convergence demonstrates that our structured narrative representations capture similar predictive patterns while offering enhanced explainability relative to traditional black-box approaches. Implementation details and performance are reported in App. \ref{app:exp-models}.

\begin{figure}[t!]
    \centering
    \includegraphics[width=0.8\columnwidth]{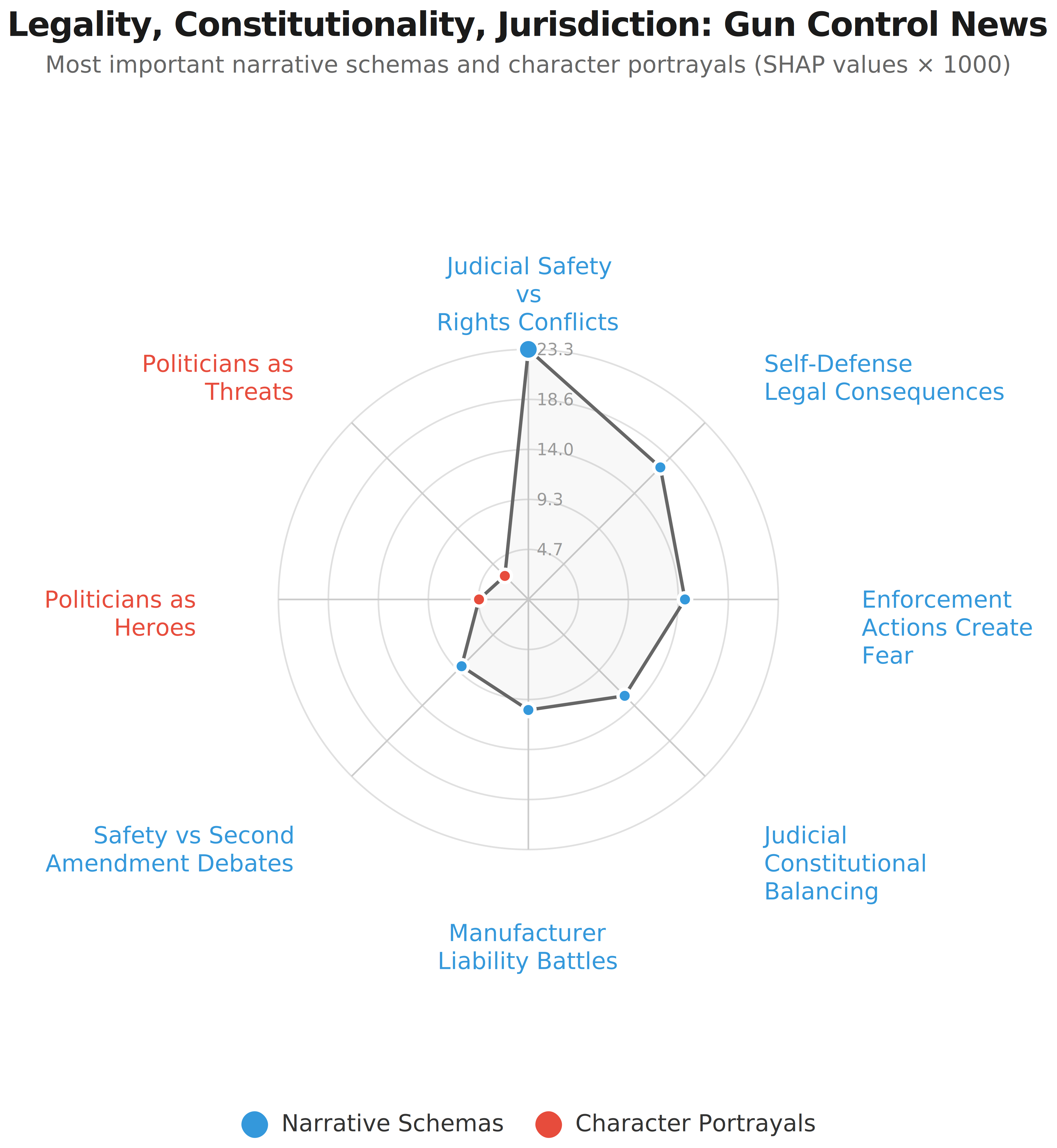}
    \caption{SHAP feature importance values (×1000) for the top 6 narrative schemas and top 2 role features that predict the Legality, Constitutionality, Jurisdiction frame in gun control news articles. Full length narrative schemas are listed in App. \ref{app:gc-legal-themes}.}
    \label{fig:gc_legal_frame}
\end{figure}

\paragraph{Frame-Level Analysis.} To move from a fine-grained (instance-level) to a medium-grained (frame-level) analysis, we compute SHAP feature importance values for each label in the dataset. This analysis allows us to understand the distinctive narrative signatures that characterize different framing strategies in a given domain. For example, Fig. \ref{fig:gc_legal_frame} illustrates SHAP values for the \textit{Legality, Constitutionality, Jurisdiction} frame in gun control coverage, showing that judicial conflict schemas seemingly dominate while political actors remain peripheral. This result suggests that the legal framing of gun control coverage tends to emphasize courtroom battles rather than legislative politics. Such frame-specific decompositions illuminate the compositional logic underlying the formation of media frames, showing how different combinations of narrative elements produce systematically different representations of the same policy issues. See more analysis in App. \ref{app:exp-frame-level}.

\paragraph{Domain-Level Analysis.} Finally, we perform a coarse-grained analysis at the domain-level to examine the most important features across all frames. This global analysis reveals the fundamental narrative building blocks that most reliably distinguish between different framing strategies within each policy domain. We find that for immigration, characters play a central role in driving frame prediction, particularly immigrants. In contrast, gun control frame predictions are driven by narrative clusters, particularly those dealing with judicial and legal schemas. See App. \ref{app:exp-domain} for full results.

\section{Conclusion}

This work addresses a fundamental challenge in computational media framing analysis: the need for methods that can automatically induce nuanced narrative schemas from large-scale, domain-agnostic corpora while maintaining interpretative depth. We propose a structured clustering framework that bridges communication theory and NLP by extracting causal narrative chains, identifying character roles, and employing character-role constraints to systematically induce semantically coherent narrative schemas at scale. Our evaluation demonstrates that our schema induction approach consistently improves narrative coherence and frame prediction performance over standard clustering methods. Further, we show that our schemas capture \citealp{Entman1993FramingTC}'s framing elements and provide interpretative power comparable to contextualized embeddings while offering superior transparency. Our results demonstrate how a unified event- and character-centric approach can systematically induce high quality, domain-specific narrative patterns.
\section*{Limitations}
While our framework demonstrates effective induction of narrative schemas, several aspects present opportunities for extension. Our focus on English-language news coverage in the United States enabled robust development using established language tools and narrative theory. Porting our solution to languages that are less supported could present challenges. However, adapting our method to other languages and contexts opens opportunities for cross-cultural analysis to explore how different societies construct meaning around similar policy issues, potentially revealing universal versus culture-specific storytelling patterns. 

The Hero/Threat/Victim framework offers a straightforward, theoretically grounded approach to character analysis, although limited. Our tractable system creates pathways for more sophisticated character modeling that could capture evolving characterizations, mixed representations, or context-dependent portrayals across different coverage contexts.

In addition, our analysis treats articles as static snapshots, which limits our understanding of how narratives develop and shift over time in response to evolving events or policy changes. Future extensions could incorporate temporal modeling that could examine how schemas emerge, persist, or fade over different time periods.

Finally, the simple classification approaches we employed prioritize explainability over performance, ensuring that the induced schemas remain accessible to communication researchers. However, we acknowledge that current architectures could provide much better performance for tasks such as frame prediction. We believe that this foundation will enable future exploration of more sophisticated computational architectures while preserving interpretative clarity.

\section*{Ethical Considerations}
Our framework's reliance on LLMs for both character role annotation and narrative schema identification raises concerns about bias amplification through probabilistic pattern reinforcement. LLMs make predictions based on statistical regularities in training data, which can systematically favor dominant cultural perspectives over marginal viewpoints when assigning character archetypes and identifying prevalent narrative schemas. This probabilistic bias toward statistically common patterns may result in certain demographic groups being consistently categorized in specific roles while simultaneously amplifying mainstream narrative framings at the expense of alternative or minority perspectives. The large-scale nature of our analysis could potentially compound this issue, as both dominant character portrayals and prevalent narrative structures become further entrenched through systematic identification across thousands of documents. 

To mitigate these risks, we recommend conducting bias audits across demographic groups and narrative schemas, developing techniques to surface potential annotation biases, and encouraging the use of this framework primarily for academic research and media literacy purposes rather than applications that could exploit induced schemas for manipulation or propaganda.

\section*{Acknowledgments}

We thank Dananjay Srinivas, Aaron Gluck, and other members of the BLAST group at the University of Colorado Boulder for the rich discussions and feedback. We thank the anonymous reviewers for their constructive feedback. This work utilized the Blanca condo computing resource at the University of Colorado Boulder. Blanca is jointly funded by computing users and the University of Colorado Boulder.

\appendix
\section{Character Group Identification Details} 
\label{app:char_ext_details}
Here we outline the process for determining salient character groups for a given corpus. This includes a prompting technique to extract character mentions from each document, clustering these character mentions into semantically-similar groups, then manually labeling the resulting clusters to obtain our final list of character groups.

\subsection{Character Extraction}

\newcommand{\specialcell}[2][l]{%
 \begin{tabular}[#1]{@{}l@{}}#2\end{tabular}}

  \begin{table*}[ht]
    \centering
    \resizebox{\textwidth}{!}{%
    \begin{tabular}{ll}
        \toprule
        \textbf{Character Group} & \textbf{Extracted Character Examples} \\
        \midrule
        \specialcell[c]{\textbf{Immigrants}} & 
        \specialcell[c]{Documented Immigrants, Undocumented Aliens, Undocumented Minors, Peruvian Immigrant}
        \vspace{0.3cm}\\

        \specialcell[c]{\textbf{Refugees}} & 
        \specialcell[c]{Asylees, Immigrant Refugee, Palastinian Refugees, Haitian Refugee Community}
        \vspace{0.3cm}\\

        \specialcell[c]{\textbf{Asylum Seekers}} & 
        \specialcell[c]{Asylum Applicants, Asylum Seekers, Chinese Asylum Seekers, Haitian Asylum Seekers}
        \vspace{0.3cm}\\

        \specialcell[c]{\textbf{Workers}} & 
        \specialcell[c]{Agricultural Guest Workers, Illegal Workers, Assembly Workers, Tech Workers}
        \vspace{0.3cm}\\

        \specialcell[c]{\textbf{Politicians}} & 
        \specialcell[c]{Donald Trump, Barack Obama, Republican Political Leadership, Republican Gov. Jan Brewer}
        \vspace{0.3cm}\\

        \specialcell[c]{\textbf{Law Enforcement}} & 
        \specialcell[c]{Alameda Police, City Police Chief, Immigration Police, Customs Agents, U.S. Border Patrol}
        \vspace{0.3cm}\\

        \specialcell[c]{\textbf{Judiciary}} & 
        \specialcell[c]{Department of Justice, Federal District Court Judges, U.S. District Judge Alan Gold, Family Court Judge}
        \vspace{0.3cm}\\

        \specialcell[c]{\textbf{Government}} & 
        \specialcell[c]{Government Workers, Department of Health and Human Services, Egyptian Government, Department of City Planning}
        \vspace{0.3cm}\\

        \specialcell[c]{\textbf{Immigration Advocates}} & 
        \specialcell[c]{Pro-Immigration Group, Latino Civil Rights Group, Immigration Rights Group, Advocacy Group}
        \vspace{0.3cm}\\

        \specialcell[c]{\textbf{Other}} & 
        \specialcell[c]{U.S. Tech Firms, Baby Boomers, Big Business, Black Market}
        \vspace{0.3cm}\\

        \bottomrule
    \end{tabular}
    }
    \caption{Final character groups for the immigration domain and selected examples of characters extracted from the dataset. For our analysis, we merge Refugess, Asylum Seekers, and Workers into the Immigrants group.}
    \label{tab:imm_char_groups}
\end{table*}

  \begin{table*}[ht]
    \centering
    \resizebox{\textwidth}{!}{%
    \begin{tabular}{ll}
        \toprule
        \textbf{Character Group} & \textbf{Extracted Character Examples} \\
        \midrule
        \specialcell[c]{\textbf{Politicians}} & 
        \specialcell[c]{Republican State Senator, Texas Gov. George W. Bush, Joe Biden}
        \vspace{0.3cm}\\

        \specialcell[c]{\textbf{Gun Control Advocates}} & 
        \specialcell[c]{Advocates of Gun Control, Anti-gun Movement, Anti-Gun-Violence Activists, Colorado Coalition \\Against Gun Violence}
        \vspace{0.3cm}\\

        \specialcell[c]{\textbf{Gun Right Advocates}} & 
        \specialcell[c]{Gun Rights Activists, Gun Lobbyists, Pro-Gun Organizations, Rocky Mountain Gun Owners}
        \vspace{0.3cm}\\

        \specialcell[c]{\textbf{Law Enforcement}} & 
        \specialcell[c]{Local Law Enforcement Agencies, D.C. Police Chief, Airport Security Officials, Hempstead Police Department}
        \vspace{0.3cm}\\

        \specialcell[c]{\textbf{Judiciary}} & 
        \specialcell[c]{California Supreme Court, Federal Judge, District Court Judge, U.S. District Judge Phyllis Hamilton}
        \vspace{0.3cm}\\

        \specialcell[c]{\textbf{Government}} & 
        \specialcell[c]{City Government, Federal Government, Governing Bodies, Government Agencies}
        \vspace{0.3cm}\\

        \specialcell[c]{\textbf{Gun Crime Victims}} & 
        \specialcell[c]{Victims of Gun Violence, Mass-Shooting Victims, Relatives of Victims, Sandy Hook Victims}
        \vspace{0.3cm}\\

        \specialcell[c]{\textbf{Other}} & 
        \specialcell[c]{Academic Experts, Adoption Agencies, Commentator, Air Line Pilots Association}
        \vspace{0.3cm}\\

        \bottomrule
    \end{tabular}
    }
    \caption{Final character groups for the gun control domain and selected examples of characters extracted from the dataset.}
    \label{tab:gc_char_groups}
\end{table*}

We prompt the Llama3.3 70B-instruct model \citep{Dubey2024TheL3} with temperature set to 0.2 in a five-shot setting. The five examples were randomly selected from each dataset and manually annotated by two of the authors. The same examples are used for each news article.

We first provide a system prompt that explains the task and desired output. This is followed by a user prompt, where the full news article is provided. The user prompt is repeated for each of the five examples. Prompts are provided in App. \ref{box:char_ext_prompt}.

\subsection{Character Clustering Details}
Given the extracted characters from all articles, we use a seeded clustering process to obtain groups of related characters. First, we compile a list of expected character groups to be used as the initial centers of our clusters.  While domain knowledge may assist in this process, it is sufficient to determine these by manually examining the raw characters extracted by prompting.

\paragraph{Immigration Seeds} undocumented immigrants, documented immigrants, law enforcement, families, politicians, voters, activists, researchers, journalists, business owners

\paragraph{Gun Control Seeds} gun owners, gun control supporters, law enforcement, politicians, voters, activists, researchers, journalists, business owners \\

The expected groups served as the initial centers for the k-means clustering algorithm. We remove all duplicates from the list of extracted characters and use scikit-learn’s \texttt{TfidfVectorizer} to map each to a corresponding Term Frequency-Inverse Document Frequency (TF-IDF) vector. These vectors are used to cluster the character mentions into semantically-similar groups. We use scikit-learn's \texttt{KMeans}\footnote{\url{https://scikit-learn.org/stable/modules/generated/sklearn.cluster.KMeans.html}} implementation with default parameters, k set to 15, and seed set to 14 to obtain clusters. We manually label the resulting character clusters to obtain our final list of character groups.

\subsection{Character Group Results}
\label{app:character_group_results}
Here we outline the final character groups for each dataset, including examples from the character extraction step which fit into each character group. Results for the immigration domain are in Tab. \ref{tab:imm_char_groups}, while results for the gun control domain are in Tab. \ref{tab:gc_char_groups}.

\section{Structured Clustering Details}

\subsection{Constraint Definition}
\label{app:constraints}
For each narrative chain, we derive a canonical representation of its character and role configuration by abstracting from individual character mentions to their corresponding character groups. When a chain contains multiple characters assigned to the same character group but exhibiting conflicting roles, we employ majority voting to resolve the inconsistency and retain the dominant character-role pairing. Random selection resolves any remaining ties. If the chain has a ``Pro'' or ``Anti'' stance, it is retained (``Neutral'' is discarded). The resulting canonical form contains at most one instance of each character group and stance combination. Narrative chains with conflicting character-role configurations are assigned cannot-link constraints, indicating that they should not be clustered together. Similarly, chains with no character-role configuration conflicts maybe linked together, but we do not explicitly enforce these may-link constraints in our clustering algorithm. Examples of cannot-link and may-link constraints are provided in Tab. \ref{tab:constraints_examples}.

\subsection{Algorithm}
\label{app:cluster_algo}

The structured clustering algorithm is outlined in Alg. \ref{alg:constrained-kmeans}.

\begin{algorithm}[t]
\caption{Constrained k-means Clustering}
\label{alg:constrained-kmeans}
\begin{algorithmic}[1]
\STATE \textbf{Input:} $\mathcal{X} = \{x_i\}_{i=1}^n$ (SBERT embeddings of narrative chains), \\
       \hspace{3em} $\mathcal{C}$ (cannot-link constraints), \\
       \hspace{3em} $k$ (number of clusters), \\
       \hspace{3em} $w_{c}$ (constraint weight)
\STATE \textbf{Output:} Partitioning $\{\mathcal{X}_h\}_{h=1}^k$
\STATE \textbf{Method:}
\STATE Initialize $\{\mu_h^{(0)}\}_{h=1}^k$ via constraint-aware \\
       \hspace{1em} k-means++
\STATE $t \leftarrow 0$
\WHILE{not converged}
    \FOR{each $x_i \in \mathcal{X}$}
        \STATE Assign cluster:
        \STATE $h^* \leftarrow \arg\min_h \big[ \|x_i - \mu_h^{(t)}\|^2$ \\
        \hspace{3em} $+ w_{c} \sum_{(x_i,x_j) \in \mathcal{C}} \mathbf{1}[l_j^{(t)} = h] \big]$
        \STATE Assign $x_i$ to cluster $h^*$
    \ENDFOR
    \STATE Re-estimate centroids:
    \STATE $\mu_h^{(t+1)} \leftarrow \frac{1}{|\mathcal{X}_h^{(t+1)}|} \sum_{x_i \in \mathcal{X}_h^{(t+1)}} x_i$ \\
           \hspace{1em} for each cluster $h$
    \STATE $t \leftarrow t + 1$
\ENDWHILE
\end{algorithmic}
\end{algorithm}

\section{Narrative Schema Attribution Details}
\subsection{Narrative Chain Sampling Strategy}
\label{app:cluster_sampling_strategy}
The system employs a two-stage sampling strategy: (1) centroid-based filtering that retains the 50\% of chains most proximate to cluster centroids via Euclidean distance, followed by (2) document-diversity sampling that prioritizes inter-document representation by first selecting one chain per unique document before additional random sampling, with a maximum of 25 chains sampled per cluster.


\section{Dataset Details}
\label{app:dataset}
\subsection{Framing Dimensions}
The Media Frames Corpus has article level frame labels derived from the \citep{Boydstun2014TrackingTD} Policy Framework, viz. \textit{Economic}, \textit{Capacity and Resources}, \textit{Morality}, \textit{Fairness and Equality}, \textit{Legality, Constitutionality and Jurisprudence}, \textit{Policy Prescription and Evaluation}, \textit{Crime and Punishment}, \textit{Security and Defense}, \textit{Health and Safety}, \textit{Quality of Life}, \textit{Cultural Identity}, \textit{Public Opinion}, \textit{Political}, \textit{External Regulation and Reputation}, \textit{Other}.

\subsection{Dataset Sampling}
We employ stratified sampling across both Immigration and Gun Control domains to preserve the original label distributions from the source datasets. The \textit{Other} label does not appear in either dataset and is discarded from our analysis. The gun control domain exhibited greater class imbalance compared to immigration. Consequently, the Capacity and Resources, Fairness and Equality, and Morality and Quality of Life labels, which had insufficient representation, were excluded from the gun control analysis. Articles with no extracted narrative chains were discarded. Final dataset statistics are reported in Tab. \ref{tab:data-stats}.

\begin{table}[ht]
\resizebox{\columnwidth}{!}{%
\begin{tabular}{@{}cccccc@{}}
\toprule
\textbf{Domain} & \textbf{Articles} & \textbf{\begin{tabular}[c]{@{}c@{}}Narrative \\ Chains\end{tabular}} & \textbf{\begin{tabular}[c]{@{}c@{}}Average Chains \\ per Article\end{tabular}} & \textbf{Constraints} & \textbf{Frame Labels} \\ \midrule
Immigration     & 1988              & 36,956                                                               & 23                                                                             & $\sim$33.3 million   & 14                    \\
Gun Control     & 1933              & 44,269                                                               & 28                                                                             & $\sim$37.4 million   & 10                    \\ \bottomrule
\end{tabular}%
}
\caption{Final sampled dataset statistics for both Immigration and Gun Control domains.}
\label{tab:data-stats}
\end{table}

\section{Evaluation Details}

\subsection{Causality Evaluation Details}
\label{app:causal-eval}


In addition to the target domain development set, we also report the performance of the teacher model on popular general-domain ECI datasets: Event StoryLine Corpus (ESC) and MavenERE ~\cite{caselli-vossen-2017-event, wang-etal-2022-maven} (Tab. \ref{tab:teacher-eci}). As ESC and MavenERE only mark causal relations, we sample equal number of non-causal relations from respective datasets.

\begin{table}[ht]
\resizebox{\columnwidth}{!}{%
\begin{tabular}{llll}
\hline
\textbf{Dataset}                            & \textbf{Precision}     & \textbf{Recall} & \textbf{F1} \\ \hline
EventStoryLine Corpus (ESC)                 & 55.3            & 96.5            & 70.3          \\
MavenERE                                    & 54.1            & 88.2            & 67.1          \\ \hline
\end{tabular}%
}
\caption{Performance of teacher model on popular Event Causality Identification (ECI) datasets (Dev F1)}
\label{tab:teacher-eci}
\end{table}

The DAPrompt model was trained on the teacher-annotated silver labels with five different seeds. We report statistics from different seed runs in Tab. \ref{tab:multiseed-eci}. The model was trained using Adam optimizer with a learning rate of $1e-6$, linear learning rate scheduling with warmup (20\% of the first epoch) and a weight decay of $1e-2$. We trained the model for a maximum of 150 epochs and calibrated based on performance on the development set. Silver labels improved performance of the system by $\approx 24 \,$ points over training on general-domain data (ESC and Maven). 

\begin{table}[ht!]
\resizebox{\columnwidth}{!}{%
\begin{tabular}{llll}
\hline
\textbf{Seed}                          & \textbf{Precision}     & \textbf{Recall} & \textbf{F1} \\ \hline
7                                     & 50.3            & 78.9            & 61.4          \\
14                                   & 46.5            & 75.6            & 57.6          \\ 
21                                   & 48.9            & 75.6            & 59.4          \\ 
42                                   & 42.1            & 82.9            & 55.9          \\ 
63                                   & 46.7            & 76.4            & 58.0          \\ \hline
Mean ($\mu$)                                   & 46.9            & 77.9            & 58.5          \\ \hline
Standard Deviation ($\sigma$)                                   & 3.1            & 3.1            & 2.1          \\ \hline
\end{tabular}%
}
\caption{Multi-seed evaluation of the DAPrompt ECI setup on the dev set}
\label{tab:multiseed-eci}
\end{table}

\subsection{Cluster Purity Details}

\paragraph{Mathematical Formulation.}
\label{app:purity-math}

Let $N$ denote the total number of narrative chains, $K$ the number of clusters, $R$ the number of distinct character groups, and $V = \{Hero, Threat, Victim\}$. For each cluster $i \in {1, 2, \ldots, K}$, we define $|C_i|$ as the cardinality of cluster $i$.

The exact match purity is defined as:

\begin{equation}
    \text{Exact Match Purity} = \frac{1}{N} \sum_{i=1}^{K} \max_{c} \text{count}(c, C_i)
\end{equation}

where $c$ represents a complete canonical configuration and $\text{count}(c, C_i)$ denotes the frequency of configuration $c$ within cluster $i$. 

For role-based purity, we compute the purity for each character type $r$ as:

\begin{equation}
    \text{Purity}_r = \frac{1}{K} \sum_{i=1}^{K} \frac{\max_{v} n_{r,v,i}}{|C_i|}
\end{equation}

where $n_{r,v,i}$ represents the count of character $r \in R$ with role $v \in V$ in cluster $i$. The average role-based purity is obtained by averaging across all character types: 

\begin{equation}
    \text{Average Role Purity} = \frac{1}{R} \sum_{r=1}^{R} \text{Purity}_r
\end{equation}

\paragraph{Role Based Purity Report.}
\label{app:role-purity}
We report the role based purity scores for the Immigration domain in Tab. \ref{tab:immig-purity-1} and \ref{tab:immig-purity-2}, as well as those for the Gun Control domain in Tab. \ref{tab:gc-purity-1} and \ref{tab:gc-purity-2}.

\subsection{Feature Vector for Logistic Regression Model}
\label{app:feature-vector}
We construct document-level feature representations by leveraging the cluster assignments of constituent narrative chains. For a given document $d$, we first map each narrative chain to its corresponding cluster assignment. We define $f_k$ as the raw frequency of the $k$-th narrative cluster within document $d$, formally expressed as:

\begin{equation}
f_k = n_k
\end{equation}

where $n_k$ denotes the count of narrative chains assigned to the $k$-th cluster. To account for potential scale differences across clusters and ensure feature stability, we apply standardization to obtain the normalized frequency $\tilde{f}_k$ for each cluster:

\begin{equation}
\tilde{f}_k = \frac{f_k - \mu}{\sigma}
\end{equation}

where $\mu$ represents the mean frequency across all clusters and $\sigma$ denotes the corresponding standard deviation. The final document representation is constructed as a feature vector $F$ encompassing the standardized frequencies of all narrative clusters:

\begin{equation}
F = [\tilde{f}_1, \tilde{f}_2, \ldots, \tilde{f}_k]
\end{equation}

This vector encoding enables downstream classification tasks by capturing the distributional patterns of narrative structures within each document.

\subsection{Best $K$ Selection} 
\label{app:best-k}
We conduct clustering experiments on narrative chains extracted from both immigration and gun control corpora, employing standard k-means clustering as a baseline and our proposed structured clustering algorithm. Clustering performance is evaluated across a range of cluster sizes $K \in \{50, 100, 150, 200, 250, 300, 350, 400, 450, 500\}$ using both purity metrics and predictive signal measures. Optimal hyperparameter selection is determined by identifying the configuration that maximizes the joint performance across both evaluation criteria. Through this systematic evaluation, we identify $K=150$ with $w_c = 0.01$ as optimal for the immigration domain, and $K=200$ with $w_c = 0.1$ for the gun control domain. Comprehensive results for these configurations are presented in Tab. \ref{tab:clustering-stats}. We show the robustness of our approach which outperforms the baseline across both metrics for both of the domains.

\subsection{Intrusion Test Details}
\label{app:intrusion-details}
\paragraph{Sampling Strategy.}The intrusion data generation employs a stratified sampling strategy that leverages cluster-based semantic relationships to create difficulty-graded evaluation datasets. For positive example pairs, the algorithm samples from the top 25\% of chains closest to each cluster's centroid while enforcing document diversity constraints and applying Jaccard similarity thresholds (default 0.6) to prevent near-duplicate pairs. Source clusters are selected through weighted random sampling that favors underutilized clusters, ensuring balanced representation across the semantic space. When positive pairs cannot satisfy diversity requirements within the top quartile, the algorithm expands its search across the entire cluster before falling back to proximity-based selection.

The intruder sampling strategy implements a three-tier difficulty hierarchy based on inter-cluster semantic distances, generating 50 examples each of easy, medium, and hard difficulty levels per clustering method for a total of 150 examples. Easy intruders are drawn from the top 25\% most representative points within clusters that are semantically distant (top 25\% farthest) from the source cluster, while medium-difficulty intruders originate from semantically similar clusters (top 25\% closest). Hard intruders represent the most challenging case, sampled from the single most semantically similar cluster to the source. This systematic approach creates a controlled difficulty gradient where task complexity correlates with the semantic proximity between source and intruder clusters, enabling fine-grained evaluation of clustering quality and human semantic judgment capabilities.

\paragraph{Intrusion Test Results.}
The full intrusion test results for both immigration and gun control datasets are reported in Tab. \ref{tab:intrusion-test}.

\begin{table*}[ht]
    \centering
    \resizebox{\textwidth}{!}{%
    \begin{tabular}{ll}
        \toprule
        \textbf{High Quality Schema} & \textbf{Salient Frame Elements} \\
        \midrule
        \specialcell[c]{Charlton Heston leads the NRA in opposing stricter gun-control legislation by advocating for rigorous \\prosecution of federal gun law violations.} & 
        \specialcell[c]{Issue, Evaluation, Resolution}
        \vspace{0.3cm}\\
        \specialcell[c]{Law enforcement efforts lead to the indictment and prosecution of individuals involved in immigration \\violations, highlighting immigrants as both victims and threats.} & 
        \specialcell[c]{Issue, Evaluation}
        \vspace{0.3cm}\\
        \bottomrule
    \end{tabular}
    }
    \caption{Examples of high-quality generated schemas, evaluated on their overall coherence, as well as on striking a balance between coverage of and specificity to the sentences included in the context.}
    \label{tab:theme_analysis_ex}
\end{table*}

\begin{table*}[ht]
    \centering
    \resizebox{\textwidth}{!}{%
    \begin{tabular}{ll}
        \toprule
        \textbf{Low Quality Schema} & \textbf{Comment} \\
        \midrule
        \specialcell[c]{Politicians' efforts to pass anti-crime legislation with stricter gun control \\measures face opposition from Gun Right Advocates, emphasizing public safety.} & 
        \specialcell[c]{It is unclear who is advocating for `public safety'}
        \vspace{0.3cm}\\
        \specialcell[c]{The U.S. government strengthens immigration enforcement through enhanced \\security measures and database implementation to combat illegal activities.} &
        \specialcell[c]{`Illegal activities' lacks specificity to the context \\sentences, which reference falsified documents}
        \vspace{0.3cm}\\
        \bottomrule
    \end{tabular}
    }
    \caption{Examples of low-quality generated schemas, evaluated on their overall coherence, as well as on striking a balance between coverage of and specificity to the sentences included in the context. We also include justification for the low-quality label.}
    \label{tab:lq_theme_analysis_ex}
\end{table*}

\subsection{Schema Analysis Details}
\label{app:schema-analysis}
We evaluate the automatic schema attribution process through a qualitative analysis by one of the authors of the paper, who is a computational linguist with a communications background. The analysis is aimed at evaluating the quality of generated schemas and their usefulness for studying media framing.

\subsubsection{Sampling Strategy}
We use a density-based stratified sampling approach to evaluate clustering quality across different levels of coherence. Clusters are ranked by internal density, measured as the inverse of average distance from data points to cluster centers. The ranked clusters are divided into three groups: top quartile (highest density), middle two quartiles, and bottom quartile (lowest density). We sample 15 clusters from the top quartile, 20 from the middle groups, and 15 from the bottom quartile, giving us 50 clusters total. Within each selected cluster, up to 25 narrative chains are randomly sampled for evaluation. This approach ensures consistent coverage across the full range of clustering performance while keeping the analysis manageable.

\subsubsection{Qualitative Analysis}
For the 50 sample schemas from each dataset, our annotator assigns a binary quality label. A high-quality generated schema is one that appropriately represents the content of the 25 sentences included in the context, while also achieving relatively high specificity. The schema must be clear and coherent. We find that 94\% of the schemas from the immigration set and 96\% of the schemas from the gun control set meet this criteria.

Aside from evaluating schema quality, we are also interested in determining the usefulness of the schemas for conducting computational frame analysis. Here, we lean heavily on \citet{Entman1993FramingTC}'s definition of framing, in which he describes four framing `functions' or `elements': \textit{defining problems}, \textit{diagnosing causes}, \textit{making moral judgments}, and \textit{suggesting remedies}. The presence of one or many of these elements indicates framing. 

Through iterative discussion, three of the authors familiar with media framing translated this list into three framing elements more suitable to the context of our task. We found \textit{defining problems} and \textit{diagnosing causes} operationally similar enough to subsume under a single banner: \textit{issue}. We gave \textit{making moral judgements} the shorthand \textit{evaluation} and decided on the term \textit{resolution} to avoid the `illness' metaphor that may invoked by the term `remedy'. Below we outline each of these resulting frame elements and provide explanations for how they may appear in the context of our schemas. 

\begin{enumerate}
    \item \textit{Issue}: A schema contains this frame element if it references the core debate in the given domain and its causal agents, either implicitly or explicitly.
    \item \textit{Evaluation}: A schema employs this element if it includes discussion of causal agents and their effects. The means through which causal agents achieve their effects are generally included within the scope of this element. 
    \item \textit{Resolution}: The schema contains specific suggestions proposed as solutions, separate from the causal interpretation of evaluation. 
\end{enumerate}

We find high rates of the \textit{evaluation} frame element in schemas. This prominent discussion of cause and effects aligns with the fact that the schemas are generated based on clusters of causal event chain verbalizations; it is encouraging that the generated schemas remain faithful to this emphasis.

While the \textit{evaluation} element was most prominent, we also observed a relatively high number of schemas containing all three elements simultaneously. A third trend, in which the \textit{issue} and \textit{evaluation} elements co-occurred, was also identified. 

Examples of high-quality schemas and the frame elements identified by the annotator in each may be found int Tab.~\ref{tab:theme_analysis_ex}. Low-quality schemas and justifications for \textit{why} they were labeled as such are included in Tab.~\ref{tab:lq_theme_analysis_ex}.

\subsection{Error Analysis Details}
\label{app:error-analysis}

We present a detailed error analysis of our narrative extraction pipeline in Tab.~\ref{tab:error-analysis-1}. For each identified error type, we provide the original article excerpt, the extracted causal event pair, the generated verbalization, the character annotation, the assigned narrative schema, and our analysis of the error and its downstream effects.

\begin{table*}[h]
\centering
\renewcommand{\arraystretch}{1.3}
\small
\resizebox{\textwidth}{!}{%
\begin{tabular}{@{} p{1.2cm} p{4.5cm} p{2cm} p{3.8cm} p{3cm} p{3.4cm} p{4.9cm} @{}}
\toprule
\textbf{Error Type} & \textbf{Article Excerpt} & \textbf{Causal Event} \newline \textbf{Pair} & \textbf{Verbalization} & \textbf{Character Annotation} & \textbf{Assigned Schema} & \textbf{Analysis} \\ \midrule

Incorrect event extraction &
Sen. Larry Craig, R-Idaho, said Democrats would turn off voters if they used the Virginia Tech tragedy to roll back gun owners' rights:
``I could give you a litany of laws broken or lines crossed. Laws sometimes don't protect the innocent, tragically enough...'' &
((roll, tragedy), CAUSE, (roll, right)) &
The Virginia Tech tragedy prompted gun rights advocates to roll out opposition against any potential rollback of gun owners' rights. &
\textbf{Characters:} \newline
gun rights advocates (Gun Right Advocates) -- \textit{Hero} \newline
\textbf{Stance:} Anti &
Gun rights advocates heroically defend their constitutional right to bear arms against government and judicial threats, emphasizing its importance for freedom and safety. &
The correct events should be (use, tragedy) and (roll back, rights). This produces a cascading error: the causality determination and verbalization are both incorrect and unfaithful to the article. Nevertheless, the assigned schema remains faithful to the article. \\
\midrule

Incorrect causal relation prediction &
But however many people are eventually able to gain legal status, there is one thing that should be required for all seeking to make a new life in this country: 
education. &
((require, thing), CAUSE, (gain, status)) &
Requiring education for young immigrants helps them gain legal status in the U.S. &
\textbf{Characters:} \newline
young immigrants (Immigrants) -- \textit{Neutral} \newline
\textbf{Stance:} Neutral &
Immigrants face obstacles accessing education due to legal status and policy restrictions,
with politicians influencing their access through supportive or restrictive legislation. &
The causality here is subtle, if present at all. The article argues that education should be a legal requirement for immigrants seeking legal status. However, the LLM instead reasons that education ensures attainment of legal status, which is unfaithful to the article. The assigned schema is consequently incorrect, even though it is closely related to the topic being discussed.
\\
\midrule

Missing entity from the verbalization &
Workers on the front lines of welfare reform, under new orders to deny food stamps to innumerable legal immigrants... are finding that translating Federal law into reality is not coming easily.Field workers are befuddled about what to do... and some are fighting their torment about the justness of it all. &
((deny, stamp), CAUSE, (fight, torment)) &
The new rules denying food stamps to legal immigrants have caused significant emotional distress,
leading many to fight against the torment of losing essential benefits. &
\textbf{Characters:} \newline
legal immigrants (Immigrants) -- \textit{Victim} \newline
\textbf{Stance:} Pro &
The denial of food stamps to legal immigrants due to welfare reform has led to high hunger levels, prompting advocacy groups to urge Congress to restore benefits. &
Although factually plausible, the verbalization incorrectly omits the ``Workers'' entity. It is the workers who experience the ``torment',' not the immigrants. The broader article discusses the plight of immigrants and the uncertainties they face, which explains the LLM's misattribution. The assigned schema correctly addresses the hunger crisis but does not directly correspond to the event pair under consideration.
\\
\midrule

Unfaithful verbalization &
As testimony continued Monday in Sami Al-Arian's bail hearing, the question of whether he will get out of jail before trial became even murkier.
Federal immigration officials have notified the lawyers for Al-Arian and one of the other defendants in the case of their intent to keep the two men in custody,
even if the magistrate grants them bail. &
((notify, lawyer), CAUSE, (grant, bail)) &
Federal immigration officials notified Al-Arian's lawyers of their intent to keep him in custody,
which could hinder the magistrate judge from granting bail. &
\textbf{Characters:} \newline
Federal immigration officials (Law Enforcement) -- \textit{Threat} \newline
Al-Arian's lawyers (Immigration Advocates) -- \textit{Hero} \newline
\textbf{Stance:} Neutral &
Judges often act as heroes by making rulings that protect immigrants, who are frequently portrayed as victims, 
while government actions sometimes pose threats, leading to legal resolutions that address these challenges. &
The verbalization incorrectly inverts the causal relationship between the events as described in the article excerpt. The assigned schema, however, is largely faithful to the article, correctly capturing the roles played by the different entities involved in such legal proceedings.
\\
\midrule

Incorrect character annotation &
Over the past few years, the Palm Beach County School District has hired 71 teachers from around the world. 
.... Other international recruits include 49 who hail from the Philippines, ....
But if the school board approves the plan tonight, the district will start offering its small but growing crop of foreign teaching recruits a chance for permanent residency ... &
((include, 49), CAUSE, (offer, crop)) &
The Palm Beach County School District's inclusion of 49 teachers from the Philippines has led to their decision to offer these foreign educators
a chance for permanent residency as an incentive to retain them. &
\textbf{Characters:} \newline
Palm Beach County School District (Government) -- \textit{Neutral} \newline
teachers from the Philippines (Immigrants) -- \textit{Hero} \newline
\textbf{Stance:} Pro &
Efforts by educators and government officials to support immigrant students' educational needs despite enrollment challenges. &
Arguably, the School District is the hero in this narrative by offering permanent residency to its foreign hires. The assigned schema, however, is incorrect, as the article discusses immigrant teachers rather than immigrant students. Since our algorithm treats immigrant teachers and students as the same character group, it is understandable why this schema was assigned based on semantic similarity alone.

\\ \bottomrule
\end{tabular}}
\caption{Error analysis of the narrative extraction pipeline. Each row illustrates a distinct error type with the original article excerpt, extracted causal event pair, generated verbalization, character annotation, assigned narrative schema, and our analysis.}
\label{tab:error-analysis-1}
\end{table*}

\section{Explainability Analysis Details}

\subsection{Models}
\label{app:exp-models}
Here we define the model architecture and the training paradigm for the models under consideration for the explainability analysis.

\paragraph{RoBERTa Classifier.} We implemented a RoBERTa-based classifier for frame prediction using a single linear classification head. We only use the full article text as input. The model was trained using the Adam optimizer with a learning rate of $1e-5$, linear learning rate scheduling with warm-up, and weight decay of $0.3$. Training proceeded for a maximum of $35$ epochs with early stopping, and performance was evaluated on a 20\% held-out development set.

\begin{table}[ht]
\resizebox{\columnwidth}{!}{%
\begin{tabular}{lllll}
\hline
\textbf{Domain}              & \textbf{Model}     & \textbf{Train F1} & \textbf{Dev F1} & \textbf{Test F1} \\ \hline
\multirow{2}{*}{Immigration} & RoBERTa            & 69.30             & 62.70           & 61.40            \\ \cline{2-2}
                             & Narrative Features & 65.97             & 43.46           & 38.57            \\
\multirow{2}{*}{Gun Control} & RoBERTa            & 75.41             & 62.98           & 64.19            \\ \cline{2-2}
                             & Narrative Features & 74.53             & 48.74           & 46.84            \\ \hline
\end{tabular}%
}
\caption{Performance of RoBERTa and Narrative Feature Classifier on the Immigration and Gun Control Datasets. We use these models for our explainability study.}
\label{tab:explain-models}
\end{table}

\paragraph{Narrative Feature Classifier.} We adapt a multi-modal fusion approach for frame classification where the model processes three distinct feature types -- cluster, role, and stance representations -- through separate transformation pathways before combining them via a fusion layer. Each pathway applies dimensionality reduction with regularization, and the concatenated representations are processed through a final classification layer. This design enables independent feature learning while facilitating cross-modal integration for improved frame prediction performance. The cluster features are constructed as outlined in App. \ref{app:feature-vector}. Role and stance features were constructed in a similar manner. The model was trained using the Adam optimizer with a learning rate of $5e-4$, linear learning rate scheduling with warm-up, and weight decay of $1e-3$. Training proceeded for a maximum of $100$ epochs with early stopping, and performance was evaluated on the same 20\% held-out development set.

\paragraph{Model Performance.} We evaluated the performance of both models on the Immigration and Gun Control datasets using a 80/20/20 train/development/test split. Results are reported in Tab. \ref{tab:explain-models}. Given the multi-label prediction setting (14 classes for immigration, 10 for gun control) and the inherently subjective nature of framing tasks, we consider the model performance satisfactory. Our primary objective centers on explainability analysis rather than performance optimization, as we seek to assess the predictive utility of our narrative feature representations rather than achieve maximum classification accuracy.

\subsection{Narrative Schemas for the Single Instance Study}
\label{app:single-instance-themes}

The complete LLM-generated narrative schemas corresponding to Fig. \ref{fig:shap-individual} are presented below in \textbf{decreasing order} of feature importance. These schemas characterize the narrative clusters containing the chains extracted from the analyzed news article.

\begin{enumerate}
    \item Immigrants face significant challenges accessing healthcare due to restrictive policies and societal fears, with varying government and advocacy responses.
    \item The failure of U.S. immigration policies and governmental actions leads to significant hardships for immigrants, portraying them as victims without effective resolution.
    \item Immigrants facing exploitation and significant challenges receive support from advocates who push for solutions like legal status.
    \item Immigrants face significant challenges accessing essential services due to language barriers, despite some positive efforts by government and advocates.
    \item Immigrants face significant challenges in achieving necessary English proficiency, which affects their ability to integrate into society and access opportunities.
    \item The U.S. government's aggressive enforcement of deportation policies targets immigrants viewed as threats, leading to widespread arrests and expulsions.
    \item Illegal immigration is causing significant problems in the U.S., such as job competition and increased crime, necessitating stronger border control and legal measures.
\end{enumerate}

\begin{figure}[ht!]
    \centering
    \includegraphics[width=1.0\columnwidth]{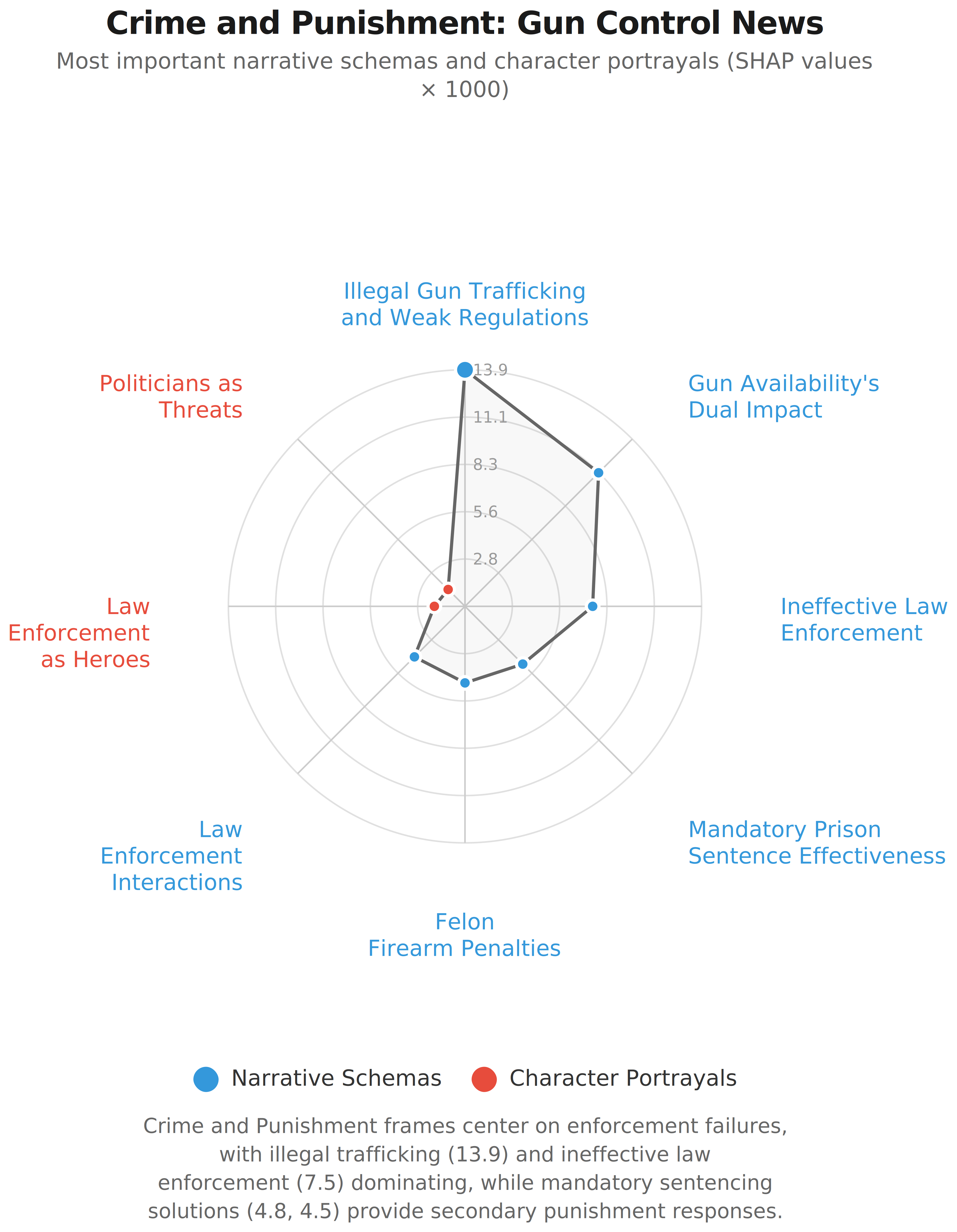}
    \caption{SHAP feature importance values (×1000) for the top 6 narrative schemas and top 2 role features that predict the Crime and Punishment frame in gun control news articles. Full length narrative schemas are listed in App. \ref{app:gc-crime-themes}.}
    \label{fig:gc_crime}
\end{figure}

\subsection{Narrative Schemas for Frame Level Analysis}
\label{app:gc-legal-themes}
The complete LLM-generated narrative schemas corresponding to Fig. \ref{fig:gc_legal_frame} are presented below in \textbf{decreasing order} of feature importance.

\begin{enumerate}
    \item Judicial decisions on gun laws spark conflicts between public safety advocates and constitutional rights supporters, shaping enforcement effectiveness.
    \item The conflict between self-defense rights and strict gun control laws leads to legal consequences for homeowners, prompting legislative changes to protect those using firearms in self-defense.
    \item The judiciary plays a central role in shaping gun control through legal rulings that balance public safety with constitutional rights.
    \item The right to bear arms is upheld as a fundamental freedom, yet its exercise leads to tragic consequences and public safety concerns.
    \item Legal battles over gun manufacturer liability drive efforts by advocates and governments to hold companies accountable, while facing opposition from gun rights supporters and politicians aiming to protect the industry.
    \item The conflict between public safety concerns and Second Amendment rights drives debates over gun control measures.
\end{enumerate}

\subsection{Additional Frame Level Analysis}
\label{app:exp-frame-level}
\subsubsection{Crime and Punishment - Gun Control}
\label{app:gc-crime-themes}
The complete LLM-generated narrative schemas corresponding to Fig. \ref{fig:gc_crime} are presented below in \textbf{decreasing order} of feature importance.

\begin{enumerate}
    \item Illegal gun trafficking, facilitated by weak regulations and loopholes, leads to increased crime, prompting law enforcement actions and calls for stricter legislation.
    \item The impact of gun availability on public safety is highlighted through incidents of self-defense and tragic shootings, illustrating both protection and loss.
    \item The ineffective enforcement of existing gun control laws leads to continued criminal access to firearms, necessitating stricter regulations and enhanced oversight.
    \item Strict enforcement of mandatory prison sentences for illegal firearm possession and sales effectively reduces gun-related crimes through law enforcement and political support.
    \item Stricter penalties for felons possessing or using firearms effectively reduce gun crimes through mandatory prison terms and deterrence.
    \item The consequences of gun possession and laws lead to interactions with law enforcement and varying outcomes. 
\end{enumerate}

\subsubsection{Cultural Identity - Immigration}
\label{app:immig-culture-themes}
\begin{figure}[ht!]
    \centering
    \includegraphics[width=1.0\columnwidth]{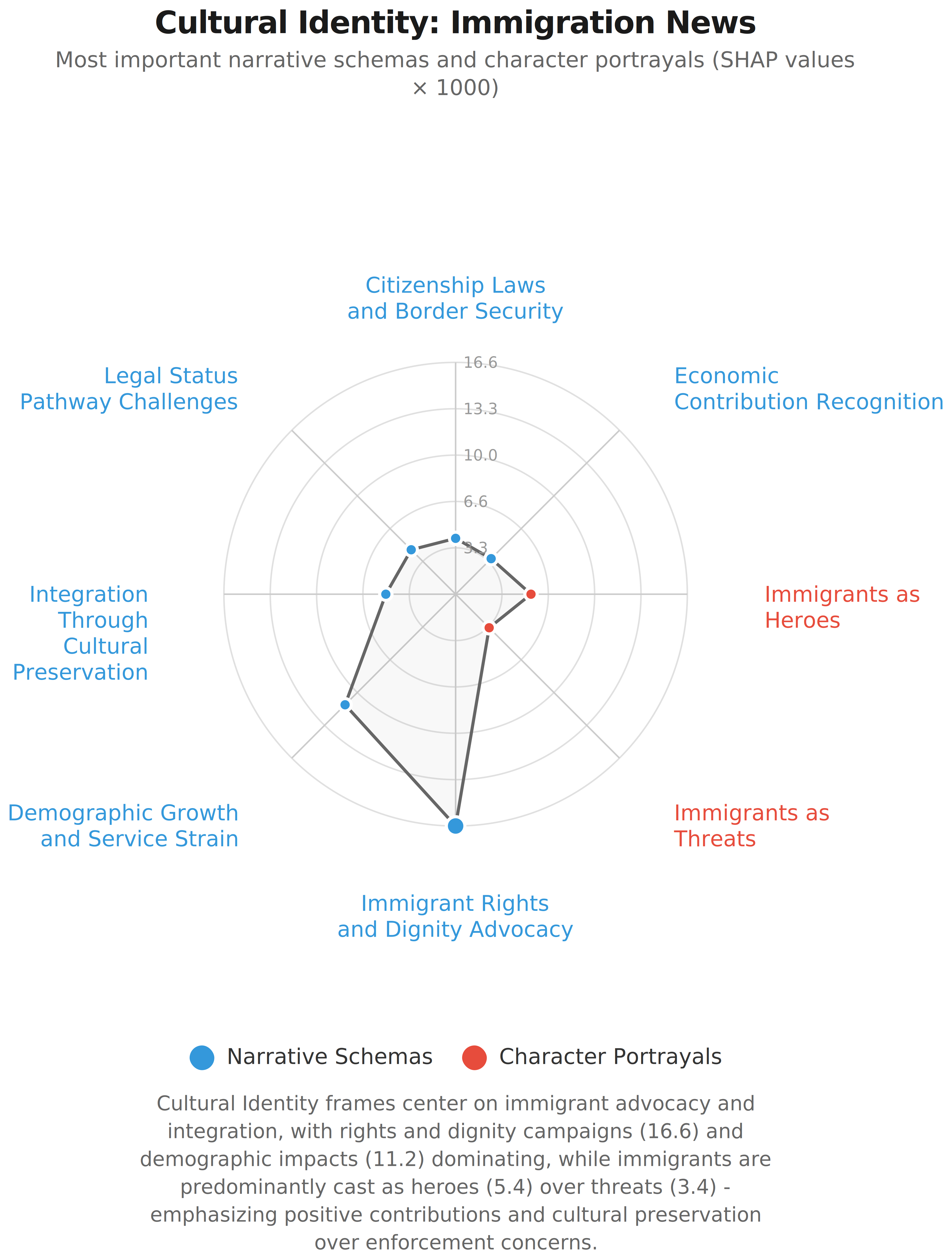}
    \caption{SHAP feature importance values (×1000) for the top 6 narrative schemas and top 2 role features that predict the Cultural Identity frame in immigration news articles. Full length narrative schemas are listed in App. \ref{app:immig-culture-themes}.}
    \label{fig:immig_culture}
\end{figure}

The complete LLM-generated narrative schemas corresponding to Fig. \ref{fig:immig_culture} are presented below in \textbf{decreasing order} of feature importance.

\begin{enumerate}
    \item Immigrants and advocates champion their rights and dignity against anti-immigrant policies, emphasizing human rights and contributions to society while advocating for reform.
    \item The rapid growth of immigrant populations is causing significant demographic changes and straining local social services.
    \item Immigrants, supported by heroic advocates like Sister Jean Marshall, overcome integration challenges through cultural preservation and language education. 
    \item Immigrants face challenges in obtaining legal status while advocates and policymakers implement measures to provide pathways for residency and integration.
    \item Politicians and advocates push for laws providing a path to citizenship for undocumented immigrants while enforcing border security to address illegal immigration fairly. 
    \item Immigrants positively contribute to the economy through work, job creation, and entrepreneurship, supported by advocates and policies. 
\end{enumerate}

\subsubsection{Economic - Immigration}
\label{app:immig-economic-themes}
\begin{figure}[ht!]
    \centering
    \includegraphics[width=1.0\columnwidth]{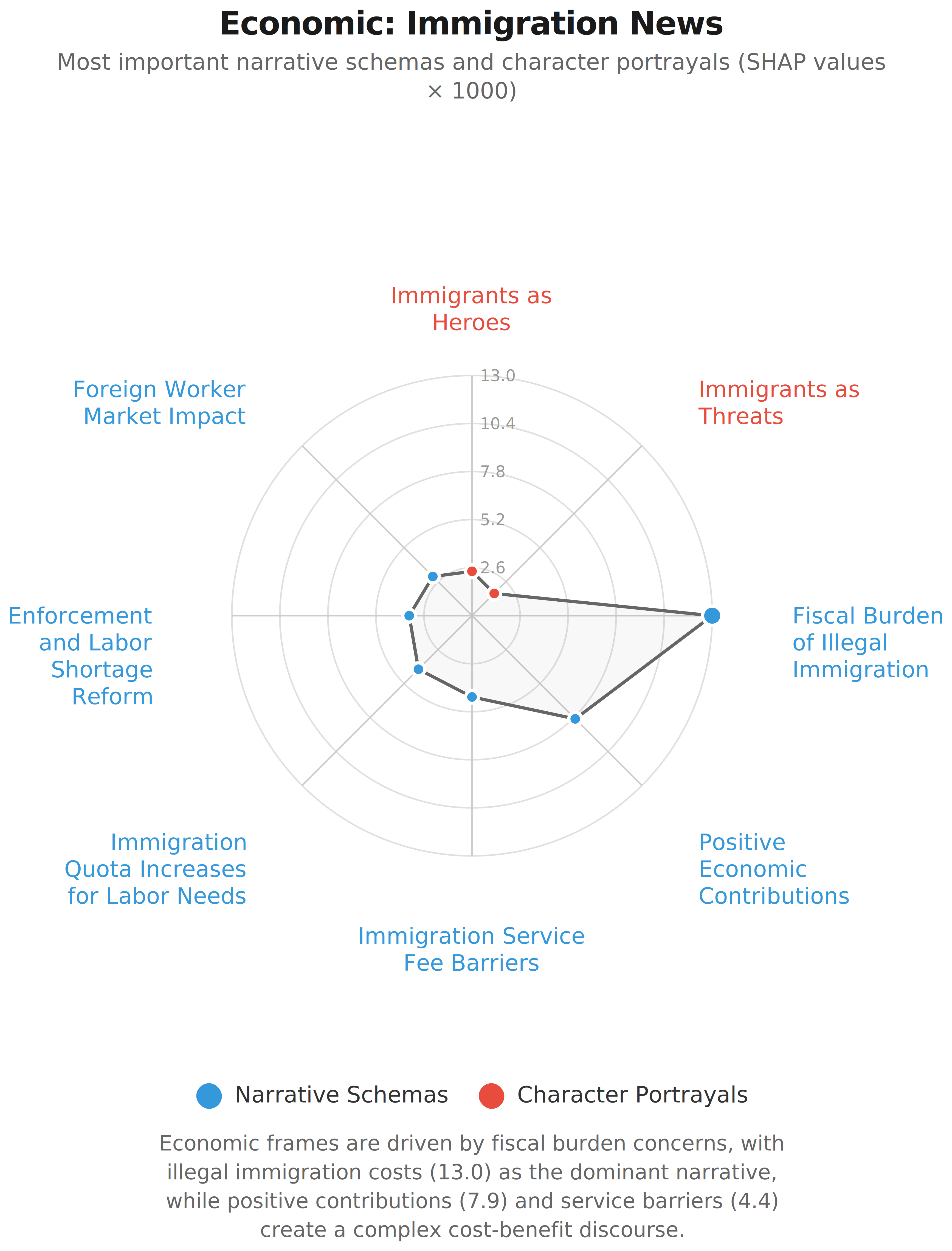}
    \caption{SHAP feature importance values (×1000) for the top 6 narrative schemas and top 2 role features that predict the Economic Identity frame in immigration news articles. Full length narrative schemas are listed in App. \ref{app:immig-economic-themes}.}
    \label{fig:immig_econ}
\end{figure}

The complete LLM-generated narrative schemas corresponding to Fig. \ref{fig:immig_econ} are presented below in \textbf{decreasing order} of feature importance.

\begin{enumerate}
    \item The high economic costs of illegal immigration impose a significant fiscal burden on federal and state governments, necessitating policy changes to reduce expenses.
    \item Immigrants positively contribute to the economy through work, job creation, and entrepreneurship, supported by advocates and policies.
    \item The increase in immigration service fees creates significant financial barriers for immigrants seeking citizenship and document services. 
    \item Increasing immigration quotas addresses labor shortages and supports economic stability through policymakers' actions.
    \item Stricter immigration enforcement and labor shortages in agriculture prompt calls for reform to stabilize the workforce. 
    \item The hiring of foreign workers impacts the U.S. job market by sometimes displacing local workers while also preventing overseas job relocation.
\end{enumerate}

\subsection{Domain Level Analysis}
\label{app:exp-domain}

\subsubsection{Immigration}
\label{app:immig-domain-themes}

\begin{figure}[ht!]
    \centering
    \includegraphics[width=1.0\columnwidth]{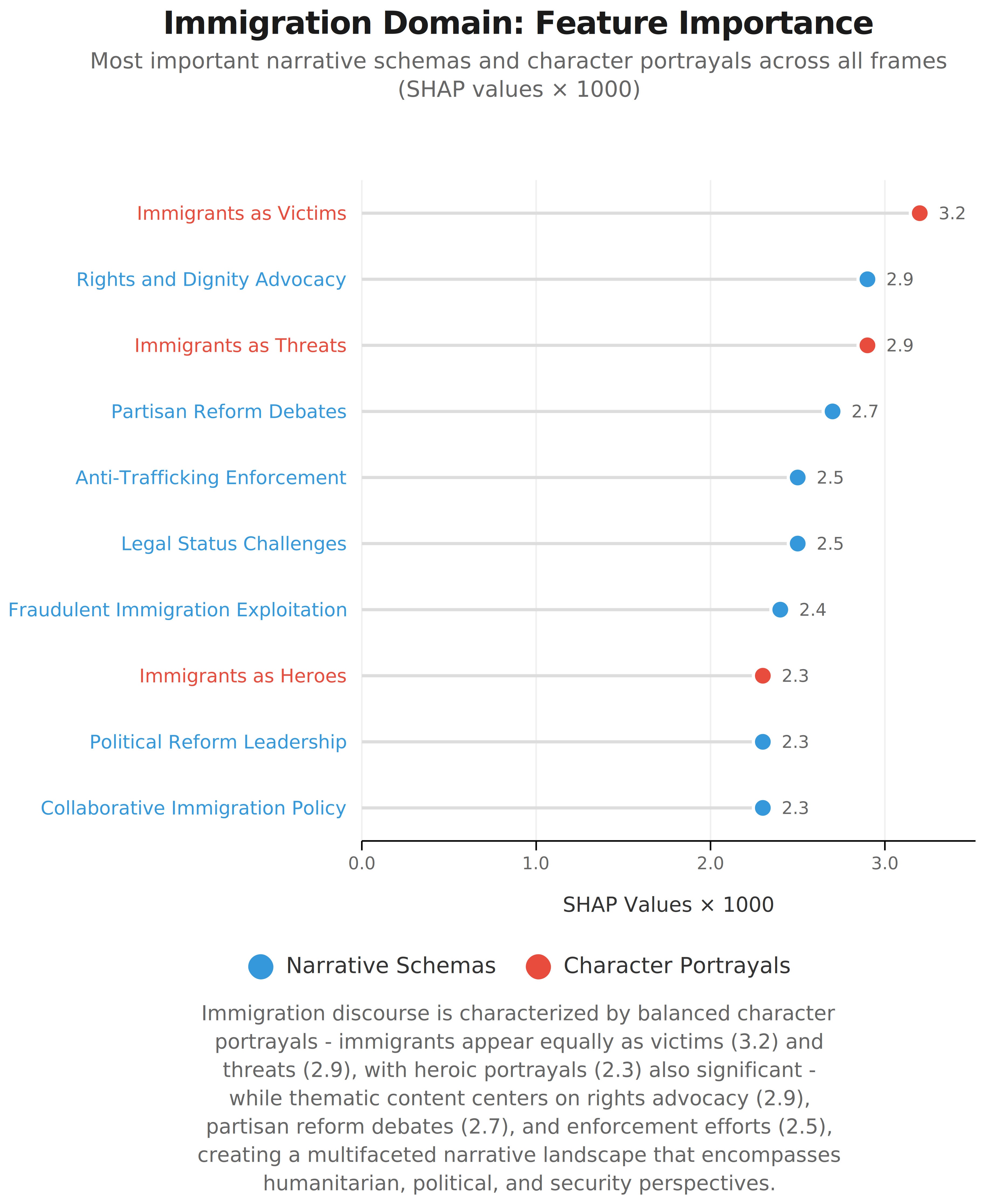}
    \caption{Top 10 most important features for frame prediction in the immigration domain, showing SHAP importance values (×1000) for narrative schemas (blue) and character portrayals (red). Features are ranked by their average contribution across all frame classifications in immigration news articles. Full length narrative schemas are listed in App. \ref{app:immig-domain-themes}.}
    \label{fig:immig_domain}
\end{figure}

The complete LLM-generated narrative schemas corresponding to Fig. \ref{fig:immig_domain} are presented below in \textbf{decreasing order} of feature importance.

\begin{figure}[ht!]
    \centering
    \includegraphics[width=1.0\columnwidth]{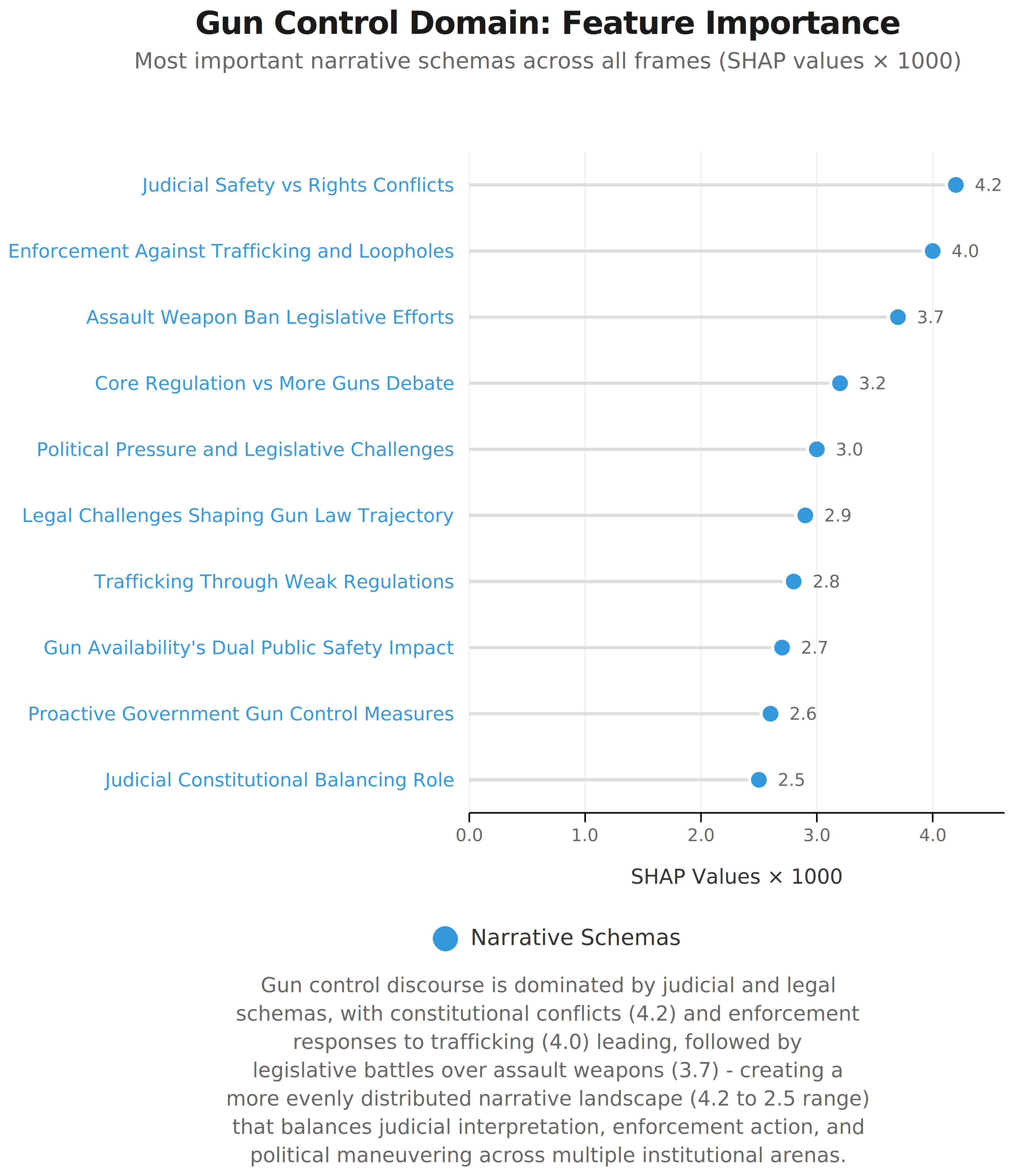}
    \caption{Top 10 most important features for frame prediction in the immigration domain, showing SHAP importance values (×1000) for narrative schemas (blue) and character portrayals (red). Features are ranked by their average contribution across all frame classifications in gun control news articles. Full length narrative schemas are listed in App. \ref{app:gc-domain-themes}.}
    \label{fig:gc_domain}
\end{figure}

\begin{enumerate}
    \item Immigrants and advocates champion their rights and dignity against anti-immigrant policies, emphasizing human rights and contributions to society while advocating for reform.
    \item The contentious political debate over immigration reform highlights legislative conflicts and compromised resolutions between Democrats and Republicans.
    \item The U.S. government and law enforcement are combating illegal immigration and human trafficking, which victimize immigrants and pose threats to national security.
    \item Immigrants face challenges in obtaining legal status while advocates and policymakers implement measures to provide pathways for residency and integration.
    \item Fraudulent immigration activities exploit vulnerable individuals, prompting legal actions and advocacy efforts to combat exploitation.
    \item Politicians are leading efforts to reform immigration laws, providing pathways to legal status and citizenship for undocumented immigrants.
    \item Politicians collaborating on immigration reform to address illegal immigration through policies like temporary-worker programs and legal status.
\end{enumerate}

\subsubsection{Gun Control}
\label{app:gc-domain-themes}
The complete LLM-generated narrative schemas corresponding to Fig. \ref{fig:gc_domain} are presented below in \textbf{decreasing order} of feature importance.

\begin{enumerate}
    \item Judicial decisions on gun laws spark conflicts between public safety advocates and constitutional rights supporters, shaping enforcement effectiveness.
    \item Stricter gun control laws and enforcement efforts aim to address loopholes and illegal firearm trafficking, which pose significant public safety risks.
    \item Politicians' efforts to pass assault weapons bans address the issue of reducing criminal gun violence through legislative action despite strong opposition from gun rights advocates. 
    \item The debate over gun control highlights a central conflict between efforts to reduce violence through regulation and arguments that more firearms enhance public safety.
    \item Politicians face challenges and pressures from gun rights advocates while attempting to pass or block gun control legislation, with some actions leading to significant defeats for gun control advocates. 
    \item Judicial decisions and legal challenges shape the trajectory of gun control laws through court rulings and legislative actions.
    \item Illegal gun trafficking, facilitated by weak regulations and loopholes, leads to increased crime, prompting law enforcement actions and calls for stricter legislation.
    \item The impact of gun availability on public safety is highlighted through incidents of self-defense and tragic shootings, illustrating both protection and loss.
    \item Government and political leaders are proactively advancing gun control measures through legislative actions and policy signings.
    \item The judiciary plays a central role in shaping gun control through legal rulings that balance public safety with constitutional rights.
\end{enumerate}

\section{LLM Hyperparameters}
We used the following hyperparamters for all DeepSeek-R1-Distill-Llama-70B inference tasks: random seed = 42, temperature = 0 (for deterministic outputs), and context length = 14336.

\section{Compute Resources}
LLM inference was run on Nvidia H100 and A100 GPUs. Clustering experiments and small model training were conducted on either CPU or a single Nvidia A100. End-to-end pipeline execution required approximately 10 days for our dataset scale.

\section{Structured LLM Outputs}
For LLM inference tasks requiring JSON-formatted outputs, schema compliance was enforced through a validation pipeline utilizing a smaller Gemma 3 4B model \citep{Kamath2025Gemma3T} for efficient output verification and reformatting.

\section{Characteristics of Annotators}
All annotators (N=4) were graduate students, aged 22-32 years, with balanced gender representation (2 male, 2 female). The team comprised two Computer Science PhD students (both experienced in media framing analysis), one Computer Science Master's student, and one Linguistics Master's student with a background in communications theory.

\section{Use of Pre-Existing Artifacts}
All pre-existing artifacts utilized in this study, including datasets, software libraries, models, and computational tools, are publicly available under open-source and open-access licenses. This academic work adheres to all intended use guidelines and terms of service for the respective resources.

\begin{table*}[ht]
\resizebox{\textwidth}{!}{%
\begin{tabular}{@{}cccccccccc@{}}
\toprule
\multirow{2}{*}{\textbf{Domain}} & \multirow{2}{*}{\textbf{Model}} & \multicolumn{2}{c}{\textbf{Government}}  & \multicolumn{2}{c}{\textbf{Immigrants}}  & \multicolumn{2}{c}{\textbf{Immigration Advocates}} & \multicolumn{2}{c}{\textbf{Judiciary}}   \\ \cmidrule(l){3-10} 
                                 &                                 & \textbf{Purity@25} & \textbf{Purity@100} & \textbf{Purity@25} & \textbf{Purity@100} & \textbf{Purity@25}      & \textbf{Purity@100}      & \textbf{Purity@25} & \textbf{Purity@100} \\ \midrule
\multirow{2}{*}{Immigration}     & k-means                 & 81.90               & 73.23               & 82.73              & 78.05               & 95.59                   & 93.02                    & 96.82              & 91.92               \\ \cmidrule(lr){2-2}
                                 & Structured Clustering                   & 83.04              & 73.08               & 85.13              & 79.55               & 94.45                   & 91.68                    & 96.74              & 92.10         \\ \bottomrule      
\end{tabular}%
}
\caption{Role-based purity scores for Government, Immigrants, Immigration Advocates, and Judiciary character groups in the Immigration domain. Metrics evaluated using all narrative chains and top 25\% chains nearest to cluster centroids.}
\label{tab:immig-purity-1}
\end{table*}

\begin{table*}[ht]
\resizebox{\textwidth}{!}{%
\begin{tabular}{@{}cccccccc@{}}
\toprule
\multirow{2}{*}{\textbf{Domain}} & \multirow{2}{*}{\textbf{Model}} & \multicolumn{2}{c}{\textbf{Law Enforcement}} & \multicolumn{2}{c}{\textbf{Politicians}} & \multicolumn{2}{c}{\textbf{Stance}}      \\ \cmidrule(l){3-8} 
                                 &                                 & \textbf{Purity@25}   & \textbf{Purity@100}   & \textbf{Purity@25} & \textbf{Purity@100} & \textbf{Purity@25} & \textbf{Purity@100} \\ \midrule
\multirow{2}{*}{Immigration}     & k-means                 & 88.42                & 80.48                 & 82.92              & 77.17               & 79.77              & 71.67               \\ \cmidrule(lr){2-2}
                                 & Structured Clustering                   & 87.53                & 79.03                 & 84.16              & 77.73               & 83.36              & 77.19 \\ \bottomrule              
\end{tabular}%
}
\caption{Role-based purity scores for Law Enforcement and Politicians character groups as well as Stance Purity in the Immigration domain. Metrics evaluated using all narrative chains and top 25\% chains nearest to cluster centroids.}
\label{tab:immig-purity-2}
\end{table*}

\begin{table*}[ht]
\resizebox{\textwidth}{!}{%
\begin{tabular}{@{}cccccccccc@{}}
\toprule
\multirow{2}{*}{\textbf{Domain}} & \multirow{2}{*}{\textbf{Model}} & \multicolumn{2}{c}{\textbf{Government}}  & \multicolumn{2}{c}{\textbf{Gun Control Advocates}} & \multicolumn{2}{c}{\textbf{Gun Crime Victims}} & \multicolumn{2}{c}{\textbf{Gun Right Advocates}} \\ \cmidrule(l){3-10} 
                                 &                                 & \textbf{Purity@25} & \textbf{Purity@100} & \textbf{Purity@25}      & \textbf{Purity@100}      & \textbf{Purity@25}    & \textbf{Purity@100}    & \textbf{Purity@25}     & \textbf{Purity@100}     \\ \midrule
\multirow{2}{*}{Gun Control}     & k-means                 & 85.23              & 77.73               & 91.37                   & 86.98                    & 95.34                 & 93.62                  & 79.57                  & 72.24                   \\ \cmidrule(lr){2-2}
                                 & Structured Clustering                   & 80.05              & 90.40                & 84.05                   & 96.45                    & 90.68                 & 84.59                  & 75.32                  & 92.10                    \\ \bottomrule
\end{tabular}%
}
\caption{Role-based purity scores for Government, Gun Control Advocates, Gun Crime Victims, and Gun Right Advocates character groups in the Gun Control domain. Metrics evaluated using all narrative chains and top 25\% chains nearest to cluster centroids.}
\label{tab:gc-purity-1}
\end{table*}

\begin{table*}[ht]
\resizebox{\textwidth}{!}{%
\begin{tabular}{@{}cccccccccc@{}}
\toprule
\multirow{2}{*}{\textbf{Domain}} & \multirow{2}{*}{\textbf{Model}} & \multicolumn{2}{c}{\textbf{Judiciary}}   & \multicolumn{2}{c}{\textbf{Law Enforcement}} & \multicolumn{2}{c}{\textbf{Politicians}} & \multicolumn{2}{c}{\textbf{Stance}}      \\ \cmidrule(l){3-10} 
                                 &                                 & \textbf{Purity@25} & \textbf{Purity@100} & \textbf{Purity@25}   & \textbf{Purity@100}   & \textbf{Purity@25} & \textbf{Purity@100} & \textbf{Purity@25} & \textbf{Purity@100} \\ \midrule
\multirow{2}{*}{Gun Control}     & k-means                 & 92.84              & 88.27               & 85.66                & 80.64                 & 84.82              & 77.69               & 76.38              & 72.24               \\ \cmidrule(lr){2-2}
                                 & Structured Clustering                   & 92.33              & 89.54               & 91.35                & 82.34                 & 86.66              & 80.5                & 85.11              & 80.68               \\ \bottomrule
\end{tabular}%
}
\caption{Role-based purity scores for Judiciary, Law Enforcement, and Politicians character groups as well as Stance Purity in the Gun Control domain. Metrics evaluated using all narrative chains and top 25\% chains nearest to cluster centroids.}
\label{tab:gc-purity-2}
\end{table*}

\begin{table*}[ht]
\resizebox{\textwidth}{!}{%
\begin{tabular}{@{}cccccccccccc@{}}
\toprule
\multirow{2}{*}{\textbf{Domain}} & \multirow{2}{*}{\textbf{Model}} & \multicolumn{2}{c}{\textbf{Easy Accuracy}}  & \multicolumn{2}{c}{\textbf{Medium Accuracy}} & \multicolumn{2}{c}{\textbf{Hard Accuracy}}  & \multicolumn{2}{c}{\textbf{Overall}}        & \multicolumn{2}{c}{\textbf{\begin{tabular}[c]{@{}c@{}}Inter Annotator \\ Agreement\end{tabular}}} \\ \cmidrule(l){3-12} 
                                 &                                 & \textbf{Annotator A} & \textbf{Annotator B} & \textbf{Annotator A}  & \textbf{Annotator B} & \textbf{Annotator A} & \textbf{Annotator B} & \textbf{Annotator A} & \textbf{Annotator B} & \textbf{Cohen's kappa}           & \textbf{95\% C.I.}                             \\ \midrule
\multirow{2}{*}{Immigration}     & k-means                 & 96.00                & 98.00                & 92.00                 & 88.00                & 76.00                & 74.00                & 88.00                & 86.67                & \multirow{2}{*}{80.99*}                          & \multirow{2}{*}{{[}67.50, 94.47{]}}            \\ \cmidrule(lr){2-2}
                                 & Structured Clustering                   & 94.00                & 92.00                & 88.00                 & 92.00                & 68.00                & 60.00                & 83.33                & 81.33                &                                                  &                                                \\ \midrule
\multirow{2}{*}{Gun Control}     & k-means                 & 92.00                & 92.00                & 84.00                 & 88.00                & 50.00                & 46.00                & 75.33                & 75.33                & \multirow{2}{*}{94.96*}                          & \multirow{2}{*}{{[}87.44, 102.48{]}}           \\ \cmidrule(lr){2-2}
                                 & Structured Clustering                   & 94.00                & 92.00                & 88.00                 & 86.00                & 70.00                & 78.00                & 84.00                & 85.33                &                                                  &                                                \\ \bottomrule
\end{tabular}%
}
\caption{Intrusion task accuracy stratified by difficulty level (easy, medium, hard) comparing k-means and our algorithm across both domains. Cohen's kappa inter-rater reliability reported. *p \textless 0.001}
\label{tab:intrusion-test}
\end{table*}

\section{Prompts}
\label{app:prompts}
We include all of the prompts used in our framework starting from App. \ref{box:causal_prompt} and onward.

\clearpage

\renewcommand{\thetextbox}{\thesection.\arabic{textbox}}

\begin{textbox*}[htbp]
\centering
\begin{tcolorbox}[
    colback=lightgreen,            
    colframe=darkgreen,            
    width=\textwidth,              
    arc=3mm,                       
    boxrule=1pt,                   
    left=5mm,                      
    right=5mm,                     
    top=3mm,                       
    bottom=3mm                     
]
{\ttfamily\small
\textbf{System Prompt} \\
\\
\# Event Causality Identification \\
\\
You are an expert annotator. Your task is to determine causality between event pairs in a news article. Events correspond to what we perceive around us and are denoted as a (VERB, OBJECT) tuple. The object is the direct object of the verb in a linguistic sense. An example of an event is (arrest, people). The verb and object will correspond to words in the article and may or may not be in their lemmatized form. There is a causal relationship between a pair of events if either EVENT\_1 causes EVENT\_2 or EVENT\_1 is caused by EVENT\_2. Directionality of the causal relationship does not matter. \\
\\
\#\# Annotation Guidelines\\
\\
- Focus only on the context provided; do not make assumptions based on world knowledge not present in the text. \\
- Consider both explicit causal markers (e.g., "because", "led to") and implicit causation. \\
- If the context is insufficient to determine causality, indicate this in your justification. \\
- Justify your decision by stating your reasoning briefly. Also, your reasoning should not simply say that "EVENT\_1" caused "EVENT\_2." or "There is a causal relationship between EVENT\_1 and EVENT\_2.". \\
\\
\#\# Input Format \\
\\
For each potential causal event pair, you will receive: \\
- DOMAIN: 'Immigration' or 'Gun Control' \\
- EVENT\_1: (VERB\_1, OBJECT\_1) \\
- SENTENCE\_1: Sentence in which EVENT\_1 appears \\
- EVENT\_2: (VERB\_2, OBJECT\_2) \\
- SENTENCE\_2: Sentence in which EVENT\_2 appears \\
- ARTICLE: Full article in which the events appear \\
\\
\#\# Output Format \\
\\
Provide your answer in this structured format: \\
\textasciigrave \textasciigrave \textasciigrave json \\
\{ \\
\phantom{....}"reason": "your reasoning for the answer", \\
\phantom{....}"relation": "causal/none" \\
\} \\
\textasciigrave \textasciigrave \textasciigrave
\\
}
\end{tcolorbox}
\captionof{textbox}{Causal Relation Extraction (contd. on next page)}
\label{box:causal_prompt}
\end{textbox*}

\begin{textbox*}[htbp]
\centering
\begin{tcolorbox}[
    colback=lightgreen,            
    colframe=darkgreen,            
    width=\textwidth,              
    arc=3mm,                       
    boxrule=1pt,                   
    left=5mm,                      
    right=5mm,                     
    top=3mm,                       
    bottom=3mm                     
]
{\ttfamily\small
\textbf{User Prompt} \\
\\
DOMAIN: <Domain of the document> \\
EVENT\_1: <The first event> \\
SENTENCE\_1: <Sentence in which EVENT\_1 appears> \\
EVENT\_2: <The second event> \\
SENTENCE\_2: <Sentence in which EVENT\_2 appears> \\
ARTICLE: <Full text of the news article>
}
\end{tcolorbox}
\caption{Causal Relation Extraction (contd.)}
\label{box:causal_prompt_ctd}
\end{textbox*}

\begin{textbox*}[htbp]
\centering
\begin{tcolorbox}[
    colback=lightgreen,            
    colframe=darkgreen,            
    width=\textwidth,              
    arc=3mm,                       
    boxrule=1pt,                   
    left=5mm,                      
    right=5mm,                     
    top=3mm,                       
    bottom=3mm                     
]
{\ttfamily\small
\textbf{System Prompt} \\
\\
\#\# Task Overview \\
You will generate a concise, plausible sentence that expands on a causal event chain formatted as verb-object pairs from a news article, incorporating relevant characters and reflecting the causal relationship accurately. \\
\\
\#\# Understanding Event Chains\\
- Events correspond to what we perceive around us and are denoted as a (VERB, OBJECT) pair \\
- The OBJECT is the direct object of the VERB in a linguistic sense \\
- The verb and object will correspond to words in the article and may or may not be in their lemmatized form \\
- An example of an event is (arrest, people) \\
- An event chain comprises two events connected by a causal relation, denoted as: $(EVENT_1, CAUSAL, EVENT_2)$ \\
- CAUSAL indicates that either $EVENT_1$ caused $EVENT_2$ or $EVENT_2$ caused $EVENT_1$ \\
- Example chain: $((arrest, people), CAUSAL, (protest, legislation))$ \\
\\
\#\# Generation Guidelines \\
- Create a single, concise sentence that clearly expresses the causal relationship \\
- Include all elements from the event chain \\
- Incorporate relevant characters/organizations from the CHARACTER GROUPS that appear in the article \\
- Maintain the article's context and factual alignment \\
- Focus only on the specific causal relationship in the event chain \\
- Do not add information or characters not relevant to the event chain \\
- Keep the sentence natural and journalistic in tone \\

\#\# Input Format \\
Each generation task will include: \\
- **DOMAIN**: The topic area (Gun Control or Immigration) \\
- **EVENT CHAIN**: A representation of two causally connected events in the format $((VERB_1, OBJECT_1), CAUSAL, (VERB_2, OBJECT_2))$ \\
- **CHARACTER GROUPS**: List of predefined character categories relevant to the domain \\
- **ARTICLE**: The complete news article containing the events \\
\\
\#\# Output Format \\
Provide your generated sentence in this JSON structure: \\
\textasciigrave \textasciigrave \textasciigrave json \\
\{ \\
\phantom{....}"sentence": "Your sentence that expands on the event chain here" \\
\} \\
\textasciigrave \textasciigrave \textasciigrave
\\
\textbf{User Prompt} \\
\\
DOMAIN: <Domain of the document> \\
EVENT CHAIN: <The corresponding narrative chain triple> \\
CHARACTER GROUPS: <Relevant character groups> \\
ARTICLE: <Full text of the news article>
}
\end{tcolorbox}
\begin{center}
\captionof{textbox}{Causal Narrative Chain Verbalization}
\label{box:chain_verb_prompt}
\end{center}
\end{textbox*}

\begin{textbox*}[htbp]
\centering
\begin{tcolorbox}[
    colback=lightgreen,            
    colframe=darkgreen,            
    width=\textwidth,              
    arc=3mm,                       
    boxrule=1pt,                   
    left=5mm,                      
    right=5mm,                     
    top=3mm,                       
    bottom=3mm                     
]
{\ttfamily\small
\textbf{System Prompt} \\
\\
Instruction: You are an annotator who is developing a dataset for the task of identifying main characters in a news article. You will include both proper and improper nouns. Your answer should be in JSON using the following format: {`characters': [`character0', `character1', ... ]}. Do not generate anything else. \\
\\
\textbf{User Prompt} \\ 
\\
Please list all important characters (including both proper and improper nouns) present in the below article about [DOMAIN] issues. <Full Text of
the News Article>
}
\end{tcolorbox}
\caption{Character Extraction}
\label{box:char_ext_prompt}
\end{textbox*}

\begin{textbox*}[htbp]
\centering
\begin{tcolorbox}[
    colback=lightgreen,            
    colframe=darkgreen,            
    width=\textwidth,              
    arc=3mm,                       
    boxrule=1pt,                   
    left=5mm,                      
    right=5mm,                     
    top=3mm,                       
    bottom=3mm                     
]
{\ttfamily\small
\textbf{System Prompt} \\
\\
\# Character Role Annotation System \\
\\
You are an expert annotator specializing in media framing analysis. Your task is to analyze how characters are portrayed in partisan news articles by identifying relevant character groups, classifying their roles, and determining the overall stance of specific event descriptions. \\
\\
\#\# Task Overview \\
For each submission, you will: \\
1. Identify which character groups appear in a given event chain (a sentence describing causal relationships) \\
2. Classify each character's role as Hero, Victim, Threat, or Neutral based on their portrayal \\
3. Determine if the event chain's framing indicates a pro, anti, or neutral stance toward the domain topic \\
\\
\#\# Critical Constraint \\
IMPORTANT: Analyze ONLY characters and entities explicitly mentioned in the event chain. Ignore all other characters and entities in the article. The full article is for context only. \\
\\
\#\# Annotation Guidelines \\
\\
\#\#\# Character Identification \\
\\
- Only include entities explicitly mentioned in the event chain \\
- Map the identified entities to the most appropriate character group \\
- If no character group from the provided list adequately represents the entity, assign it to 'Other'.  This includes entities that are tangentially related but don't fit well into any specific predefined category, as well as entities from completely different domains or contexts. \\
- Use the exact wording from the event chain when listing specific entities \\
- Places and locations such as countries, states, cities, streets etc. are NOT valid character entities ignore them \\
\\
\#\#\# Role Classification \\
\\
- Base classification solely on how the entity is portrayed in the event chain and surrounding context \\
- Use textual evidence to justify your classification \\
- When an entity fits multiple roles, select the most prominent one \\
\\
\#\#\# Stance Determination \\
\\
- Map stance annotation to one of Pro, Anti, or Neutral \\
- Consider linguistic cues, framing devices, and emotional tone \\
- Look for partisan language that reveals underlying bias \\
- Evaluate how the event chain portrays policy positions related to the domain \\
}
\end{tcolorbox}
\caption{Character Role Extraction (contd. on next page)}
\label{box:char_role_prompt}
\end{textbox*}

\begin{textbox*}[htbp]
\centering
\begin{tcolorbox}[
    colback=lightgreen,            
    colframe=darkgreen,            
    width=\textwidth,              
    arc=3mm,                       
    boxrule=1pt,                   
    left=5mm,                      
    right=5mm,                     
    top=3mm,                       
    bottom=3mm                     
]
{\ttfamily\small
\#\# Input Format \\
Each annotation task will include: \\
- **DOMAIN**: The topic area (Gun Control or Immigration) \\
- **EVENT CHAIN**: A sentence describing a causal relationship between events/actions \\
- **CHARACTER GROUPS**: List of predefined character categories relevant to the domain \\
- **ROLE DESCRIPTIONS**: Definitions of Hero, Victim, Threat, and Neutral roles in context \\
- **ARTICLE**: The complete news article containing the event chain \\
\\
\#\# Output Format \\
Provide your analysis in this JSON format: \\
\textasciigrave \textasciigrave \textasciigrave json \\
\{ \\
\phantom{....}"characters": [ \\
\phantom{......}\{ \\ 
\phantom{..........}"entity": "[Exact entity mentioned in the event chain]", \\
\phantom{..........}"character group": "[Identified character group from predefined list]", \\
\phantom{..........}"role": "[Hero/Victim/Threat/Neutral]" \\
\phantom{......}\} \\
\phantom{....}  ], \\
\phantom{....}"stance": "[Pro/Anti/Neutral]" \\
\} \\
\textasciigrave \textasciigrave \textasciigrave
\\
\textbf{User Prompt} \\
\\
DOMAIN: <Domain of the document> \\
EVENT CHAIN: <The corresponding narrative chain verbalization> \\
CHARACTER GROUPS: <Relevant character groups> \\
ROLE DESCRIPTIONS: <Hero/Threat/Victim descriptions in the context of the domain>
ARTICLE: <Full text of the news article>
}
\end{tcolorbox}
\caption{Character Role Extraction (contd.)}
\label{box:char_role_prompt_ctd}
\end{textbox*}

\begin{textbox*}[htbp]
\centering
\begin{tcolorbox}[
    colback=lightgreen,            
    colframe=darkgreen,            
    width=\textwidth,              
    arc=3mm,                       
    boxrule=1pt,                   
    left=5mm,                      
    right=5mm,                     
    top=3mm,                       
    bottom=3mm                     
]
{\ttfamily\small
- Hero: A hero in immigration discourse is portrayed as courageously defending fundamental values against powerful opposition. Heroes are characterized by moral conviction, personal sacrifice, and principled action in the face of resistance. Depending on partisan framing, heroes might be fighting for humanitarian protection of vulnerable migrants, defending national sovereignty and security, upholding legal immigration processes, or protecting cultural identity. Heroes are described using language of advocacy, protection, leadership, and standing up for what's "right" despite significant challenges or personal cost. \\
\\
- Victim: A victim in immigration discourse is portrayed as suffering harm, injustice, or vulnerability due to current policies or the actions of opposing groups. Victims lack agency in their suffering and are presented as deserving of sympathy or protection. Depending on partisan framing, victims might include migrants fleeing persecution, citizens affected by immigration policies, taxpayers bearing economic burdens, workers facing job competition, or communities experiencing cultural disruption. Victims are described using language of innocence, hardship, desperation, and vulnerability, with their experiences serving as emotional anchors demonstrating what's at stake in policy debates. \\
\\
- Threat: A threat in immigration discourse is portrayed as dangerous to social values, safety, economic prosperity, or rule of law. Threats are characterized by power (or growing power), harmful intent or indifference, and opposition to what the narrative frames as proper social order. Partisan perspectives differ significantly in who or what constitutes a threat: pro-immigration narratives may identify xenophobic politicians, harsh enforcement agencies, or discriminatory policies as threats; anti-immigration narratives may portray undocumented immigrants, foreign criminal organizations, immigrant workers competing for jobs, or cultural incompatibility as threats. Both perspectives frame their identified threats as undermining fundamental societal values, whether those values center on humanitarian principles and diversity or national security, economic opportunity for citizens, and cultural cohesion. Threats are described using language of danger, disruption, economic harm, or disregard for important norms or principles. \\
\\
- Neutral: A neutral entity in immigration discourse is portrayed primarily as an information source, procedural actor, or contextual element without strong moral judgment. Neutral characters fulfill functional roles and are presented in matter-of-fact terms. They might include researchers providing migration statistics, certain government agencies performing routine processing functions, courts adjudicating cases according to established law, or economic factors influencing migration patterns. Neutral entities are described using objective language focused on processes, facts, and institutional functions rather than moral qualities or emotional impact.
}
\end{tcolorbox}
\caption{Role Descriptions for Immigration}
\label{box:char_role_immig}
\end{textbox*}

\begin{textbox*}[htbp]
\centering
\begin{tcolorbox}[
    colback=lightgreen,            
    colframe=darkgreen,            
    width=\textwidth,              
    arc=3mm,                       
    boxrule=1pt,                   
    left=5mm,                      
    right=5mm,                     
    top=3mm,                       
    bottom=3mm                     
]
{\ttfamily\small
- Hero: A hero in gun control discourse is portrayed as a principled actor fighting against powerful opposition to protect what the narrative presents as fundamental values (whether safety or rights). Heroes are characterized by courage, moral clarity, and sacrifice. They are framed as understanding the "real truth" of the situation while facing resistance from misguided or malevolent forces. Heroes are often described using language of protection, defense, advocacy, leadership, and standing up for vulnerable populations. \\
\\
- Victim: A victim in gun control discourse is portrayed as someone who has suffered harm, loss, or vulnerability due to current policies, societal conditions, or the actions of opposing groups. Victims are characterized by innocence, lack of agency in their suffering, and being collateral damage in larger political battles. Different partisan perspectives highlight different victim groups: some focus on those harmed by gun violence, while others emphasize those whose self-defense capabilities, constitutional rights, or cultural practices are threatened by restrictive measures. Victims frequently serve as emotional anchors in narratives, with their experiences used to demonstrate what's at stake. \\
\\
- Threat: A threat in gun control discourse is portrayed as a dangerous entity actively working to undermine safety, rights, or societal values. Threats are characterized by power, malice or callous indifference, and ideological extremism. They are framed as either deliberately causing harm or being willfully blind to the consequences of their actions/positions. Threats can be individuals, organizations, or abstract concepts (like "government overreach" or "gun culture"). They serve as the antagonist that heroes must confront and from which victims must be protected. \\
\\
- Neutral: A neutral entity in gun control discourse is portrayed primarily as an information source, procedural actor, or contextual element without strong moral judgment or emotional framing. Neutral characters are characterized by objectivity, functional roles, and matter-of-fact presentation. They may include researchers presenting data without advocating specific policies, journalists reporting events factually, institutional processes functioning as designed, or stakeholders presented in balanced terms acknowledging legitimate concerns on multiple sides. Neutral entities often provide context, background information, or procedural details that help frame the discourse without themselves becoming central to moral narratives.
}
\end{tcolorbox}
\caption{Role Descriptions for Gun Control}
\label{box:char_role_gc}
\end{textbox*}

\begin{textbox*}[htbp]
\centering
\begin{tcolorbox}[
    colback=lightgreen,            
    colframe=darkgreen,            
    width=\textwidth,              
    arc=3mm,                       
    boxrule=1pt,                   
    left=5mm,                      
    right=5mm,                     
    top=3mm,                       
    bottom=3mm                     
]
{\ttfamily\small
\textbf{System Prompt} \\
\\
\# Narrative Cluster Analysis System \\
\\
You are a computational social scientist specializing in extracting high-level narrative themes from clustered text data. Your task is to analyze collections of sentences from narrative clusters along with character role information and stance towards the topic area (e.g., Gun Control, Immigration) to identify the central narrative theme.
\\
\#\# Instructions \\
\\
1. Analyze the provided sentences to identify the central narrative theme \\
2. Consider how the character roles interact within this narrative framework \\
3. **Conservative Framing Assessment**: Only after constructing your theme, examine whether the cluster exhibits **clear and explicit** Entman's framing elements: \\
   - **Issue/Problem**: A specific, well-defined problem that is explicitly articulated (not just implied) \\
   - **Evaluation**: Clear moral judgments or strong value assessments about causes, effects, or actors (must be unambiguous positive/negative evaluations) of the issue \\
   - **Resolution/Solution**: Explicit suggestions for remedies, courses of action, or treatment recommendations (not just general implications) for actors involved in the issue \\
4. **Important**: Only detect framing patterns when these elements are **prominently and explicitly present** in the cluster sentences. Subtle implications, vague suggestions, or weak patterns should NOT trigger framing detection. \\
5. **Theme Construction**: \\
   - Generate ONE concise sentence (15-25 words) that captures the overarching narrative theme \\
   - Include specific events, actions, or circumstances that recur across the sentences \\
   - Focus on relationship dynamics, conflict, causation, or resolution patterns \\
   - If clear framing patterns exist, ensure the theme incorporates the central issue and relevant evaluation/resolution elements \\
6. **Framing Elements Identification**: If framing patterns are detected, identify the exact substrings from your generated theme sentence that correspond to each framing element. Do not paraphrase or create new text - only use exact portions of the theme sentence. **Overlapping substrings between issue, evaluation, and resolution are permitted when necessary** (e.g., when evaluative language is embedded within the issue description or when resolution language overlaps with evaluation). \\
7. Ensure your narrative summary reflects the specific domain context provided \\
8. Output your response as valid JSON in the specified format \\

\#\# Input Format \\
Each analysis task will include: \\
- **DOMAIN**: The topic area (Gun Control or Immigration) \\
- **CHARACTER ROLES**: List of possible character roles that may appear in the cluster \\
- **CLUSTER SENTENCES**: A collection of sentences representing the narrative cluster along with their character roles \\
\\
\#\# Output Format \\
Provide your analysis in this JSON format: \\
\textasciigrave \textasciigrave \textasciigrave json \\
\{ \\
\phantom{....}"theme": "[Concise sentence summarizing the central narrative theme.]", \\
\phantom{....}"framing\_elements": \{ \\
\phantom{........}"issue": null, \\
\phantom{........}"evaluation": null, \\
\phantom{........}"resolution": null \\
\phantom{....} \} \\
\} \\
\textasciigrave \textasciigrave \textasciigrave
}
\end{tcolorbox}
\captionof{textbox}{Narrative Theme Attribution (contd. on next page)}
\label{box:theme_attr_prompt}
\end{textbox*}

\begin{textbox*}[htbp]
\centering
\begin{tcolorbox}[
    colback=lightgreen,            
    colframe=darkgreen,            
    width=\textwidth,              
    arc=3mm,                       
    boxrule=1pt,                   
    left=5mm,                      
    right=5mm,                     
    top=3mm,                       
    bottom=3mm                     
]
{\ttfamily\small
\textbf{User Prompt} \\
\\
DOMAIN: <Domain of the document> \\
CHARACTER Roles: <Relevant character roles> \\
CLUSTER SENTENCES: <A list of narrative chain verbalizations with their roles>
}
\end{tcolorbox}
\captionof{textbox}{Narrative Theme Attribution (contd.)}
\label{box:theme_attr_prompt_contd}
\end{textbox*}

\end{document}